# A Generalizable Feature Extractor for Alzheimer's-Related Brain MRI Tasks

Reza Rajabli[1,2], D. Louis Collins[1,2]

[1] McConnell Brain Imaging Centre, Montreal Neurological Institute, McGill University

[2] Department of Biomedical Engineering, McGill University

**Correspondence:** Reza Rajabli (reza.rajabli@mail.mcgill.ca)

# Abstract

When there is not enough labeled data to properly train deep learning models, transfer learning can help. In this approach, a model is first trained on a related task with plenty of data, and then its learned features are used for the target task. Although this approach works well in many fields, we still do not fully understand how effective it is in neuroimaging, especially for Alzheimer's disease research. It is also not clear if these transferred models can work on new datasets without being retrained for each specific task.

We evaluate whether a compact, supervised pretrained model can serve as a reusable foundation model for downstream neuroimaging tasks. We freeze the 7.18 million weights of a custom pseudo-fully-convolutional residual network, a 3D CNN previously trained for brain-age prediction, and adapt it to each downstream task using parameter-efficient fine-tuning with Low-Rank Adaptation (LoRA), requiring only ~1% additional trainable parameters.

We evaluate generalizability in six experiments. Adapting the model to classify cognitively normal versus Dementia on ADNI gave an AUC of 0.964 (bAcc 0.889) on held-out folds (Experiment #1). Applying that adapted model unchanged to OASIS-3, with no retraining, recalibration, or adjustment, gave an AUC of 0.871 (bAcc 0.796) on a dataset the encoder had never seen (Experiment #2). Reusing its output logit together with age and a cognitive score distinguished stable from progressing MCI with an AUC of 0.828 (bAcc 0.753) in an ADNI population excluded from its training, matching a published random forest built on engineered features (Experiment #3). Adapting the same backbone to a different classification task, predicting amyloid positivity from structural MRI, gave an AUC of 0.804, in the upper part of the published range for T1w-only specialized models (Experiment #4). Finally, the same approach estimated ICV-normalized hippocampal and white matter hypointensity volumes directly from the T1w image without explicit segmentation, with $R^2$ of 0.80 and 0.91 respectively, tasks that are normally addressed with much larger U-Net segmentation networks (Experiments #5 and #6).

A compact model supervised on brain age can therefore serve as a reusable backbone across diagnosis, prognosis, molecular pathology, and morphometry, adapting to each task with approximately 1% additional parameters and transferring to an unseen cohort without any training on that cohort. These results highlight the importance of implicit regularization strategies such as LoRA for building neuroimaging models that generalize across datasets. Our findings suggest that a carefully trained brain age model can serve as an effective foundation model for Alzheimer's related tasks, even under strict data constraints.

## Introduction

Deep learning models for brain MRI typically require large labeled datasets. However, in Alzheimer's disease (AD) research, available labeled datasets are often small. To address this limitation in other fields, researchers have developed large foundation models that are pretrained using self-supervised or contrastive learning on thousands of scans and then adapted to specific tasks. This approach is based on two main assumptions that are not often examined. The first assumption is that increasing model scale is necessary. The second is that the pretrained model can generalize to a new cohort without retraining. The common method for evaluating transfer, which involves training a new classifier on each target dataset, does not test whether the model can perform well on entirely unseen data.

Among those works reporting out-of-distribution test results, Mazher et al. [1] report an AUC of 0.81 for BrainFound on CN versus AD dementia in ADNI, the cohort it was trained on, and 0.71 to 0.77 on National Alzheimer's Coordinating Center (NACC), OASIS, and AIBL without retraining. Kaczmarek et al. [2] evaluated their SimCLR-based 3D brain MRI foundation model with linear probing, in which only a linear classifier is trained on top of the frozen features, and with full fine-tuning; BrainSegFounder [3] is adapted by fine-tuning the pretrained encoder together with a U-Net decoder for segmentation. In both cases every number reported comes from a model fitted, at least in part, on the data it is tested with. Bron et al. [4] are the exception, with and AUC of 0.94 on ADNI that is nearly maintained with AUC of 0.90 on an independent Dutch memory clinic cohort. Their model was built for that single classification task from voxel-wise grey matter maps, however, and is not a reusable encoder.

Two studies are close to ours, and both point somewhere the field has not followed. Dufumier et al. [5] pretrained a 3D DenseNet121 on approximately 10,000 healthy T1w scans using participant age as continuous proxy metadata in a contrastive loss. Barbano et al. [6] proposed AnatCL, a 3D ResNet-18 pretrained on 3,984 OpenBHB images with a weakly contrastive loss that also uses anatomical measures, evaluated by linear probing on 12 downstream tasks including CN vs. AD dementia on both ADNI and OASIS-3 datasets. For both tasks, a new classifier is fitted on each target dataset.

The most informative result in both papers is one neither of them highlights. Each compares its contrastive objective against a baseline trained with ordinary supervision to predict chronological age, and in each the baseline performs comparably to the method being proposed; in Barbano et al. [6] it achieves the highest balanced accuracy in the table for separating stable from progressing mild cognitive impairment (MCI). In both papers, however, the age-supervised model appears only as a comparator: it is never scaled up, never applied to tasks beyond classification, and never evaluated without fitting a new classifier on the target dataset. If age supervision performs this well in studies designed to advocate for something else, the open question is not whether it is competitive, but how far it can be pushed. That is the question we take up here.

We propose a different approach. Our model is compact: a 3D CNN with 7.18 million parameters, trained with supervision on brain age [7]. We never update those weights. For each new task we add roughly 1% new trainable parameters using Low-Rank Adaptation (LoRA) [8]. We proceed in six experiments. We first adapt the model to a new task, CN versus Dementia classification on ADNI (Experiment #1), and then apply that adapted model unchanged to OASIS-3, a dataset it has never seen, without any retraining, recalibration, or adjustment (Experiment #2). We next reuse its output logit, together with age and a cognitive score, to distinguish stable from progressing mild cognitive impairment (MCI) in a subset of the ADNI cohort that was excluded from training (Experiment #3). We then ask whether the same backbone can be adapted to a different kind of classification: predicting amyloid positivity from structural T1w MRI (Experiment #4). Finally, we apply the same frozen backbone model to tasks that the literature solves with much larger U-Net segmentation networks: estimating normalized hippocampal and white matter hypointensity volumes directly from the image, without any explicit segmentation step (Experiments #5 and #6).

Freezing the encoder is not only a way to save computation. Kumar et al. [9] showed that full fine-tuning can distort pretrained features: across ten datasets, full fine-tuning was about 2% more accurate than a linear layer on frozen features within distribution, and about 7% less accurate out of distribution. The cause is that the lower layers change while the head is still being learned. Minoccheri et al. [10] applied this to a U-Net transferred across institutions with only 30 labelled cases and found that low-rank methods outperformed standard fine-tuning. More broadly, a benchmark of 17 parameter-efficient methods across six medical datasets found that their advantage grows as the amount of downstream data falls [11].

There are three reasons why this is important. First, the encoder is relatively small: 7.18 million frozen parameters, plus 72,249 for each task adapter. For comparison, the encoders of the BrainSegFounder models [3] contain 19 to 27 million parameters. Adding a new task to our model therefore costs about 1% more parameters. Ours can therefore be trained and deployed on more modest hardware. Second, in the external experiment, we do not fit anything on the target cohort. Therefore, the reported performance reflects true generalizability for what accuracy a new site would achieve, rather than the results possible after collecting and labeling new data. Third, the pretraining label is chronological age, a metric which is available for almost every MRI scan acquired, while other labels (e.g., disease labels) are limited, costly to obtain, and not evenly distributed across different populations.

# Methods

## Dataset Descriptions

### Alzheimer's Disease Neuroimaging Initiative (ADNI)

ADNI is a multi-site longitudinal public–private partnership aimed at identifying biomarkers for the progression of AD. We included baseline 3D T1-weighted scans from cognitively healthy participants across the ADNI-1, ADNI-Go, ADNI-2, ADNI-3, and ADNI-4 cohorts (available at

https://ida.loni.usc.edu, accessed January 2026, see Table 1). ADNI images were acquired at multiple sites on 1.5T (ADNI-1) and 3T (ADNI-Go/2/3/4) scanners from three manufacturers (GE, Siemens, and Philips) using standardized 3D MPRAGE (or vendor-equivalent IR-SPGR) protocols. Representative parameters were TR≈2,300 ms, TE≈3 ms, TI≈900-1,000 ms, flip angle≈8–9°, and near-isotropic voxel sizes of ~1.0-1.1 × 1.0-1.1 × 1.2 mm (ADNI-1/Go/2) or 1.0 × 1.0 × 1.0 mm (ADNI-3/4) [12], [13].

## Open Access Series of Imaging Studies (OASIS-3)

OASIS-3 is an open-access resource containing longitudinal data from over 1,000 subjects collected over 30 years [14]. It serves here as an external test cohort only (Experiment #2) and contributes to no training or adaptation stage. We included 1,150 participants with a usable baseline T1w scan and a diagnosis derivable from the UDS clinical diagnosis table, of whom 908 were cognitively normal and 242 had dementia attributed to AD (available at https://www.nitrc.org, accessed October 2024; see Table 1). Scans were acquired on a 3T Siemens TIM Trio using an MPRAGE sequence (TR=2400 ms, TE=3.08 ms, TI=1000 ms, flip angle=8°, 1 $mm^3$ voxels).

**Table 1 Participant characteristics for each experiment, at the earliest visit used.**
Values are mean ± SD; APOE4+ denotes at least one ε4 allele. ADAS-Cog-13 is not available for OASIS-3.

| Experiment | Group | n | Age (y) | Female | APOE4+ | Educ. (y) | CDR-SB | ADAS13 |
|---|---|---|---|---|---|---|---|---|
| #1 (ADNI) | Cognitively Normal | 370 | 73.4 ± 6.2 | 57.8% | 27.3% | 16.6 ± 2.6 | 0.02 ± 0.12 | 8.5 ± 4.3 |
| | Dementia | 177 | 74.6 ± 7.8 | 46.9% | 71.8% | 15.6 ± 2.7 | 4.71 ± 1.78 | 30.4 ± 7.5 |
| #2 (OASIS-3) | Cognitively Normal | 908 | 68.7 ± 9.1 | 58.4% | 35.3% | 16.1 ± 2.5 | 0.06 ± 0.26 | — |
| | Dementia | 242 | 76.0 ± 7.8 | 45.9% | 63.6% | 14.6 ± 2.9 | 3.78 ± 1.86 | — |
| #3 (ADNI MCI) | Stable MCI | 155 | 72.3 ± 6.8 | 46.5% | 61.9% | 16.3 ± 2.7 | 1.36 ± 0.87 | 14.2 ± 6.2 |
| | Progressing MCI | 55 | 72.9 ± 6.6 | 43.6% | 78.2% | 16.3 ± 2.5 | 2.06 ± 0.92 | 23.2 ± 7.3 |
| #4 (ADNI) | Amyloid negative | 460 | 71.7 ± 7.1 | 49.7% | 19.4% | 16.6 ± 2.5 | 0.59 ± 1.02 | 10.2 ± 5.8 |
| | Amyloid positive | 405 | 74.8 ± 7.2 | 49.0% | 65.8% | 16.1 ± 2.7 | 2.13 ± 2.12 | 18.8 ± 10.5 |
| #5 / #6 (ADNI) | All participants | 2,092 | 73.1 ± 7.5 | 48.2% | 44.7% | 16.0 ± 2.8 | 1.47 ± 1.83 | 15.8 ± 9.6 |

## Fine-tuning

Using LoRA, we fine-tuned a custom pseudo-fully-convolutional residual network that was previously trained on a brain age prediction task. This base model was developed using a regularized and weight-consolidated training procedure to improve generalization. In LoRA, we keep the model weights frozen while injecting a low-rank weight matrix decomposition into them. This keeps the number of trainable parameters extremely low, while offering performance on par with or better than traditional fine-tuning. The base encoder of our model has 7.18 million parameters, all of which we freeze. We then remove the brain-age regression head and add a small number of LoRA adapters on top of every layer together with a randomly initialized task head, adding roughly 1% new trainable parameters (72,249 new trainable parameters in total). Our goal was to evaluate the capacity of the pretrained model on unfamiliar downstream tasks, not necessarily to beat the state of the art. Therefore, in all experiments, we used the same configuration. We did not optimize the hyperparameters for each experiment, and we kept the architecture unchanged. Since we used one hyperparameter setting for all six experiments, the results represent a conservative lower bound on accuracy. Tuning the hyperparameters for each task could improve the results. We set the LoRA rank to 2, the LoRA alpha to 4.0, and the learning rate to 1e-3.

Moreover, to make sure that our results were not due to chance, we validated the model with a cross-validation scheme. We split the data into 5 folds by participant ID, so no participant's scans appeared in more than one fold. Each fold was held out as the test set three times. Each time, the remaining 4 folds were pooled and randomly split, with a different seed each time, into 80% for training and 20% for validation. This gave us 15 trained models per experiment in total. The reported metrics are the aggregated results of each model's performance on its own holdout test set, with each subject appearing in a test set exactly 3 times. Figure 1 summarizes which data are used at each stage, and shows that no OASIS-3 scan enters pretraining or adaptation.

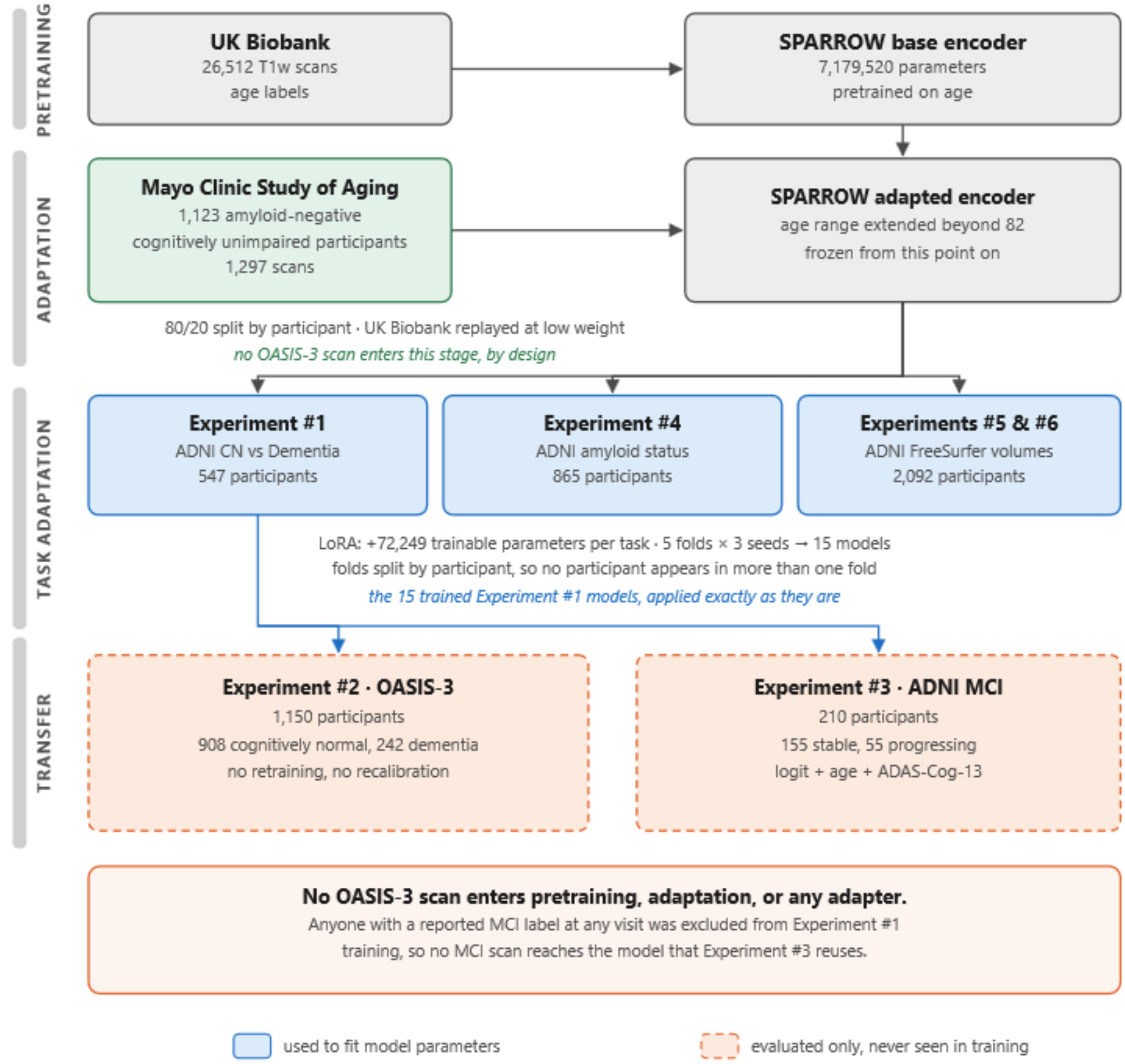

**Figure 1. Data flow through the experiments.**
First, we pretrained the encoder on UK Biobank scans using age as the target label, and extended it to the elderly range by continual learning on MCSA; the OASIS-3 adaptation cohort used in [Chapter 4] was deliberately excluded here so that OASIS-3 remains unseen. We then kept its weights fixed for all later tasks. For each new task, we added a separate set of LoRA parameters (72,249 in total) and trained them independently. Next, we tested the model from Experiment #1 on the OASIS-3 dataset without making any changes, using its output logit to tell apart stable and progressing MCI cases. It is important to note that we did not use any OASIS-3 scans during pretraining or adaptation, and there were no MCI participants in the Experiment #1 training set.

## Experiment #1: Cognitive Normal vs Dementia

While most papers in the literature call this task CN/AD classification, we call it CN vs. Dementia, since not all cases labeled as Dementia in ADNI are due to Alzheimer's disease [15]. We used the ADNI cohort to fine-tune our model here using labels from the ADNIMERGE table [16]. Diagnoses in ADNIMERGE are recorded at examination visits, which do not always coincide with imaging visits. For each participant we therefore defined two anchors: the last stable CN visit, i.e. the latest examination with a CN diagnosis such that no non-CN diagnosis appears at any earlier examination; and the first stable Dementia visit, i.e. the earliest examination with a Dementia diagnosis such that no non-Dementia diagnosis appears at any later examination. Imaging visits on or before the last stable CN visit were labelled CN, and imaging visits on or after the first stable Dementia visit were labelled Dementia. Imaging visits falling between the two anchors, and participants with fluctuating diagnoses that leave either anchor undefined, were excluded. Then, we replaced the brain age prediction head with a randomly-initialized linear head to predict a logit to classify CN vs. Dementia.

The total number of new trainable parameters is 72,249 (~1% of the brain age model's trainable parameters). The cohort comprised 547 participants. In each fold, 328 or 329 participants were used for training, contributing between 1,097 and 1,197 imaging visits, with 109 participants for validation and 109 or 110 for testing. Validation and test metrics use one visit per participant, while training uses all available visits. The training procedure was done according to what is explained in the Fine-tuning section. Participants may have several imaging visits. To use all available data without letting participants with many visits dominate training, we sampled one visit per participant at random in each training epoch, so that over training the model sees all visits but each participant contributes equally per epoch. For validation and testing we used only the earliest visit of each participant, so that those metrics are computed on one scan per participant and do not change between runs.

## Experiment #2: External Validation on OASIS-3

To test generalization to a dataset the model had never seen, we applied the Experiment #1 LoRA adapted model unchanged to the OASIS-3 dataset, with no retraining, recalibration, or adaptation of any kind. The 15 fold × seed models were combined by soft voting and evaluated on 1,150 OASIS-3 participants (908 CN, 242 with dementia), using the earliest visit per participant. Diagnostic labels for OASIS-3 were derived from the UDS clinical diagnosis table using the same stable-window logic we applied to ADNI in Experiment #1. A visit was counted as CN when NORMCOG was set and DEMENTED was not, and as dementia when NORMCOG was not set,

DEMENTED was set, and at least one Alzheimer's-specific flag was present (probable AD, possible AD, or a recorded Alzheimer's diagnosis), so that dementia not attributed to AD was excluded. For each participant we then took the last visit in the leading run of normal diagnoses and the first visit in the trailing run of dementia diagnoses. T1w scans acquired on or before the former were labelled CN, scans acquired on or after the latter were labelled dementia, and scans falling between the two, where the diagnostic history is not monotonic, were discarded. Scans were matched to diagnosis visits by the day offset recorded in the session label, reduced to one row per session, and kept only where a preprocessed T1w volume was available. One consequence should be noted: the ADNI labels used for training cover dementia of any cause, while these OASIS-3 labels are restricted to dementia attributed to AD, so the external cohort is more homogeneous than the training data. Because no OASIS-3 data was used at any stage of training or model selection, nothing in the model is fitted to this cohort.

## Experiment #3: Prediction of Future Conversion from MCI to Dementia

For the MCI-stable vs. MCI-progressive task, definitions vary across studies. We chose the definition used by Shafiee et al. [17], since we had access to the same ADNI cohort subject IDs they used for their task. This allows us to compare our results directly with theirs. Their progression was defined as a two or more increase in CDR sum of boxes (CDR-SB) score. They predicted progression over two years and achieved a bAcc of 75.9±1.1 when using age, ADAS-Cog-13 (a cognitive score [18]), and MRI as input features.

They trained a random forest for this task, using age, cognitive scores, and the SNIPE score [19] for the hippocampus (HC) and entorhinal cortex (EC). The SNIPE score shows how similar a brain structure is to a normal structure (HC or EC) or to a degenerated structure (HC or EC) in someone with dementia. To extract the SNIPE score, each structure must first be segmented for every participant, then lookup library sets of CN and dementia cases must be built, and the structure being studied compared against them. Here we ask whether a single logit from our frozen encoder can replace that feature-engineering pipeline.

Here, we instead used the output logit of the model trained on CN and Dementia participants from ADNI data in Experiment #1. First, the model was trained only on CN and Dementia participants and has seen no ADNI MCI scans, so there is no leakage into this task. Second, the output logit, which is thresholded to classify a brain as normal or demented, can be used as a severity score: the closer the logit is to 1, the more demented the brain, and the closer it is to 0, the healthier the brain.

We used the same 210 ADNI MCI subjects as Shafiee et al. and passed all the images through the model fine-tuned in Experiment #1 to obtain the output logit for each participant. We then trained a logistic regression classifier to predict future progressors over a 2-year period, using age, ADAS-Cog-13, and the extracted logit as features. We used the same fold/seed structure as before (5 folds × 3 seeds, each subject tested 3 times), but with stratified folds this time (as was done in Shafiee et al. [17]). A separate validation fold was not needed here, since logistic regression does not require early stopping. We use a two-point increase in CDR-SB as the label definition, following Shafiee et al. [17]. We set the probability threshold at 0.5 for logistic regression. The cohort has 210 subjects: 55 are progressing, and 155 are stable MCI. In each fold, we hold out 42 subjects (11 progressing and 31 stable) and train on the remaining 168. Shafiee et al. [17] also used stratified cross-validation on this cohort, but their participant splits are not the same as ours. Therefore, our comparison with their results shows how the two methods perform on the same cohort, but not on identical data splits.

## Experiment #4: Amyloid positivity

For amyloid-positivity, using only T1w images provides moderate predictive performance in the literature. For example, Lew et al. reported an AUC of 0.73 (95% confidence interval, CI: 0.68, 0.78) for predicting amyloid-positivity using only T1w brain MRIs, rising to 0.79 (95% CI: 0.74, 0.83) when MRI was combined with demographics, APOE status, cognitive scores and hippocampal volumes [20]. Chattopadhyay et al. reported a bAcc of 0.76 with a 3D CNN and an AUC of 0.857 with a vision transformer, both from T1w images alone [21]. A more recent work by Kim et al. reported an accuracy of 0.59 and an AUC of 0.61 for a model that uses only T1w images as input [22].

As stated in the introduction, identifying amyloid positivity from MR images alone is a challenging problem; however, if the frozen representation carries signal related to amyloid deposition, its performance should fall within the range reported for models that use T1w images alone (AUC 0.61–0.86, refs [20], [21], [22]). We use the amyloid-positivity labels derived by UC Berkeley from amyloid PET data in the ADNI dataset (obtained from LONI [23]), filtering out visits with amyloid-positivity fluctuations using the same approach used to retain stable-CN and stable-Dementia participants described in the Experiment #1 section. We then replace the brain-age prediction head with a randomly-initialized linear head producing a logit for binary amyloid-positivity classification, as in Experiment #1. The cohort comprised 865 participants, divided in each fold into 519 for training (contributing between 1,478 and 1,533 imaging visits), 173 for validation, and 173 for testing.

## Experiment #5: Normalized Hippocampal Volume

Most recent studies on hippocampal volume estimation rely on U-Net-based architectures with much larger networks for explicit segmentation of the hippocampus, and of the intracranial cavity for normalization. In contrast, our model predicts head-size-normalized hippocampal volume directly, without explicitly segmenting either the hippocampus or the intracranial volume (ICV). This is a more challenging prediction problem because the network must implicitly identify the image features that determine hippocampal volume while also accounting for the normalization that would conventionally require a separate estimate of ICV. In addition, normalization removes variance associated with overall head size, including signal that is correlated with hippocampal volume and could otherwise facilitate prediction. Despite these constraints, we consider normalized hippocampal volume the more clinically and scientifically relevant target, since normalized rather than raw volumes are generally used in group-level analyses.

However, if our model can estimate the hippocampus size, this shows that it was able to learn features related to the hippocampus. This means that: 1) it can work in a similar way to larger segmentation models, 2) it can focus on smaller parts of the input, or 3) there is some global structural knowledge that can be used to estimate the hippocampus size. As a result, we finetuned our model with the same approach as before to estimate the hippocampus size. For the gold standard we used hippocampal volume normalized by ICV, taken from the University of California, San Francisco (UCSF) Cross-Sectional FreeSurfer (7.x) tables, which use the standard recon-all segmentation stream, reported by UCSF (downloadable from LONI). The cohort comprised 2,092 participants, divided in each fold into 1,255 or 1,256 for training (contributing between 8,491 and 8,950 imaging visits), 418 for validation, and 418 or 419 for testing.

### Experiment #6: Normalized White Matter Hypointensity Volume

As a second continuous target, we estimated white matter hypointensity volume, normalized by ICV, directly from T1w images. White matter hypointensities are regions that appear darker than normal-appearing white matter on T1w images, and they are a commonly used proxy for the white matter lesion burden that is more often quantified as hyperintensities on T2 or FLAIR images, but correspond to more destructive lesions. Unlike hippocampal volume, this target is diffuse and variable in location, so it tests a different property of the model representation. Reference volumes were taken from the same UCSF Cross-Sectional FreeSurfer (7.x) tables distributed through LONI, which derive white matter hypointensity volumes from the same T1w images, and were normalized by ICV. We fine-tuned the frozen encoder with the same LoRA configuration and cross-validation scheme as the other experiments, replacing the brain-age head with a randomly initialized linear regression head, giving 72,249 trainable parameters as in the other experiments. This experiment used the same 2,092 participants and the same fold structure as Experiment #5.

# Results

## Experiment #1

On the held-out ADNI folds, the fine-tuned model separated CN from dementia participants with a pooled out-of-fold AUC of 0.964 (95% CI 0.947–0.978; Figure 2.1). At a 0.5 decision threshold it achieved a balanced accuracy of 0.889, a sensitivity of 0.842 (95% CI 0.787–0.893), and a specificity of 0.935 (95% CI 0.908–0.959) over 547 subjects (177 with dementia; soft-vote ensemble of the 15 fold × seed adapters), and the corresponding confusion matrix is shown in Figure 2.2. The subject-bootstrap distribution of the AUC was narrow and clearly separated from chance, as shown in Figure 2.3. The predicted probabilities were well calibrated, and the score distributions for the two groups were strongly separated (Figure 2.4). Training remained stable across all folds. The validation AUC reached a plateau, and both training and validation loss decreased together, which indicates that the model converged without overfitting (Figure 2.5).

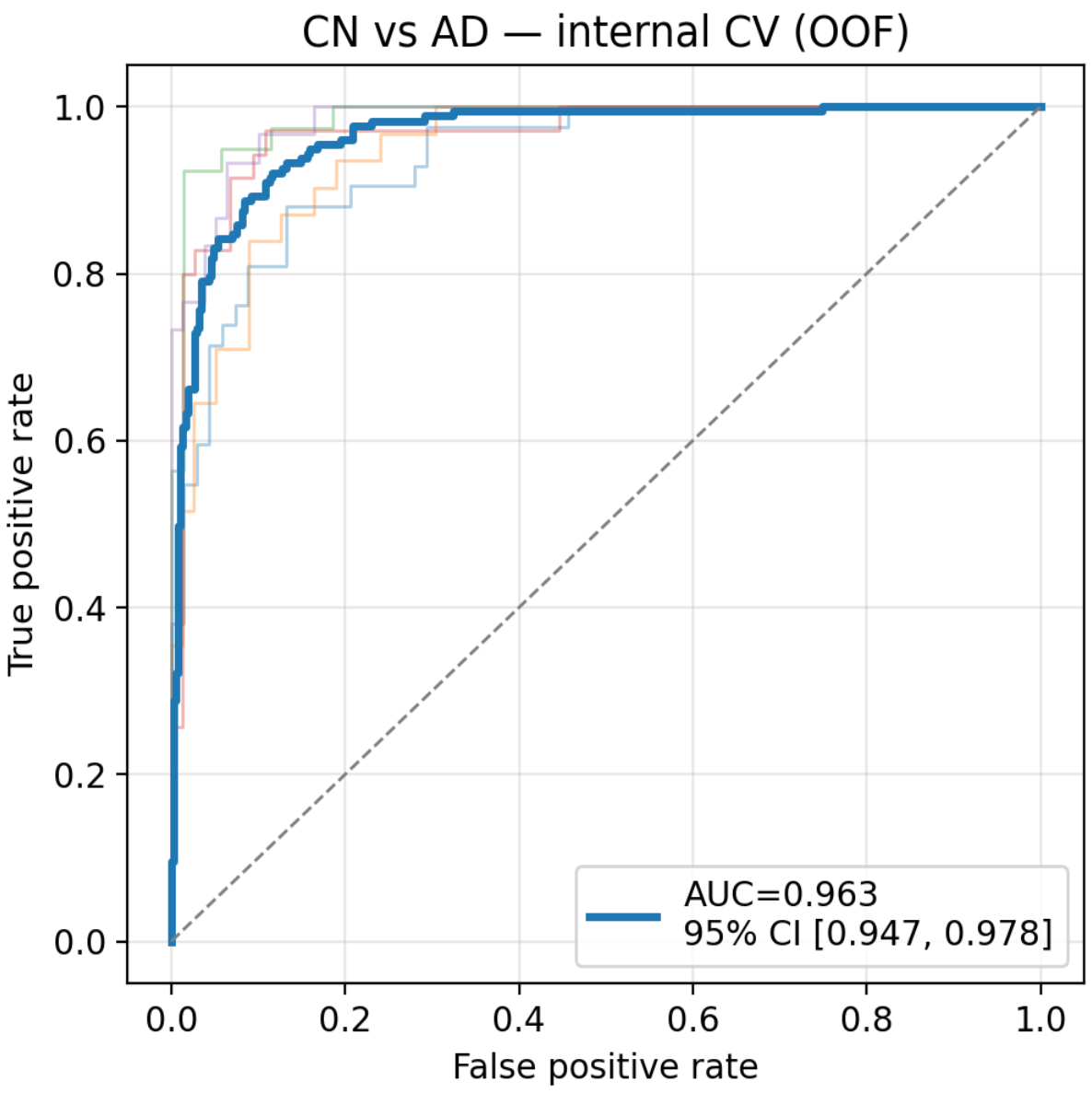

**Figure 2.1** Receiver operating characteristic (ROC) curve for CN vs Dementia classification (internal ADNI cross-validation; pooled out-of-fold AUC = 0.963 95% CI 0.947–0.978). Thin lines show the five folds; the dashed diagonal marks chance.

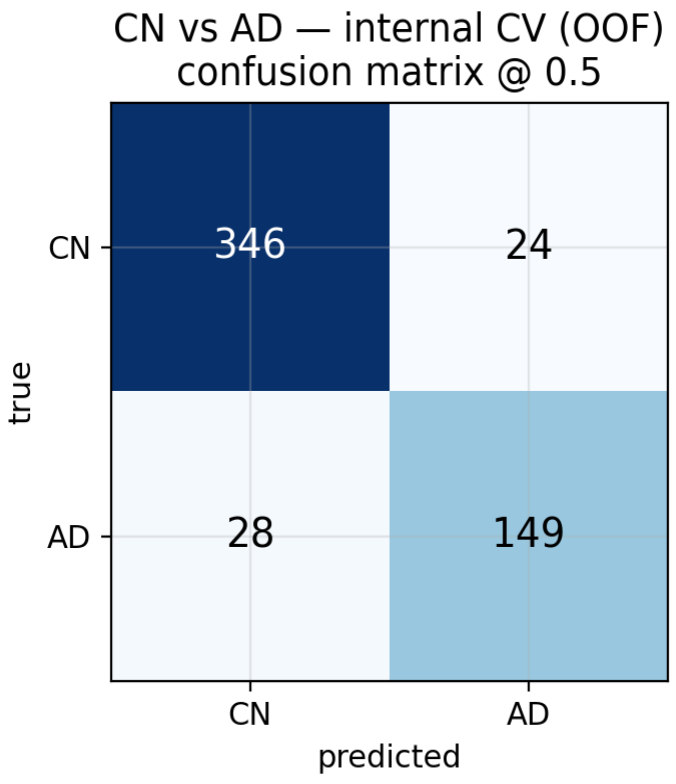

**Figure 2.2** Confusion matrix for Cognitively Normal vs Dementia classification on ADNI at a 0.5 threshold (pooled out-of-fold; 547 subjects, 177 with dementia), corresponding to a sensitivity of 0.842 and a specificity of 0.935.

**Figure 2.3** Subject-bootstrap distribution of the internal AUC (5,000 resamples). The point estimate (0.964) and the 95% confidence interval (0.947–0.978) are indicated; the distribution is narrow and well separated from chance.

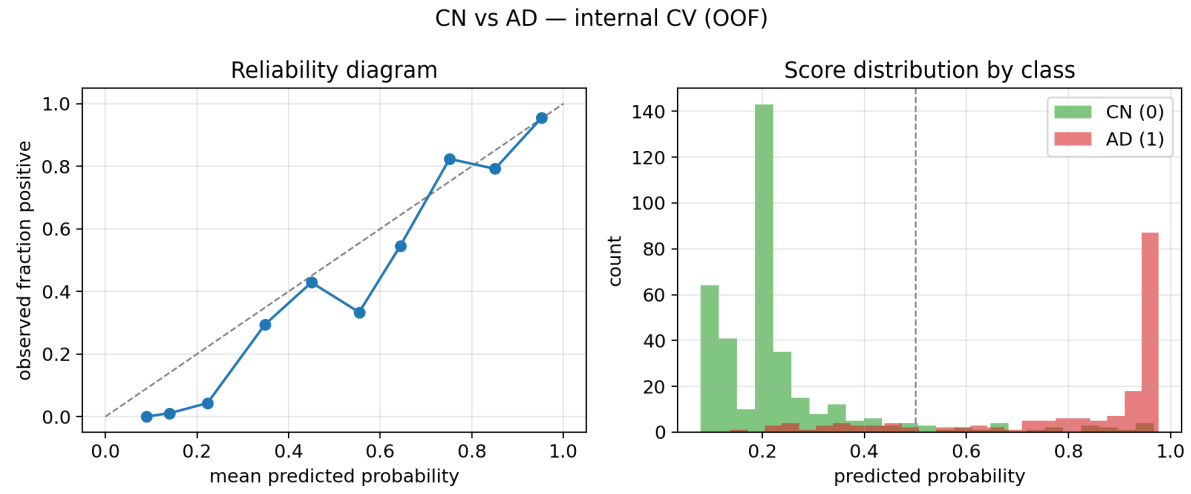

**Figure 2.4** Calibration of the internal classifier.
Left: reliability diagram (predicted vs observed probability); the curve tracks the diagonal, indicating well-calibrated probabilities. Right: predicted-probability distributions for the cognitively normal and dementia groups, showing clear separation.

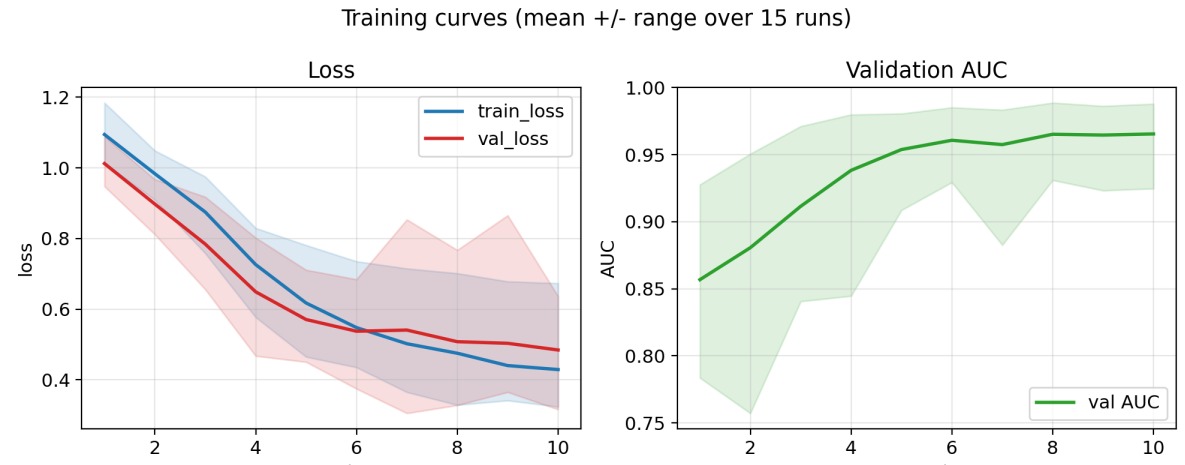

**Figure 2.5** Training dynamics across the 15 fold × seed runs (mean ± range).
Left: training and validation loss versus epoch. Right: validation AUC versus epoch, which plateaus as the losses converge, indicating stable fine-tuning without overfitting.

# Experiment #2

When we applied the model without retraining to the external OASIS-3 cohort (1,150 subjects; 908 CN and 242 with dementia), the 15-adapter ensemble still showed strong discrimination. It achieved a pooled AUC of 0.871 (95% CI 0.844–0.896; see Figure 3.1). Using a 0.5 decision threshold, the model reached a balanced accuracy of 0.796, a sensitivity of 0.727 (95% CI 0.673–0.785), and a specificity of 0.865 (95% CI 0.842–0.886). The confusion matrix is shown in Figure 3.2. The subject-bootstrap distribution of the AUC stayed clearly above chance for this new cohort (Figure 3.3), and the model remained well calibrated, with the two diagnostic groups still separated in predicted probability (Figure 3.4). The drop in AUC of about 0.09 compared to internal cross-validation (0.964 to 0.871) shows the out-of-distribution gap between ADNI and OASIS-3.

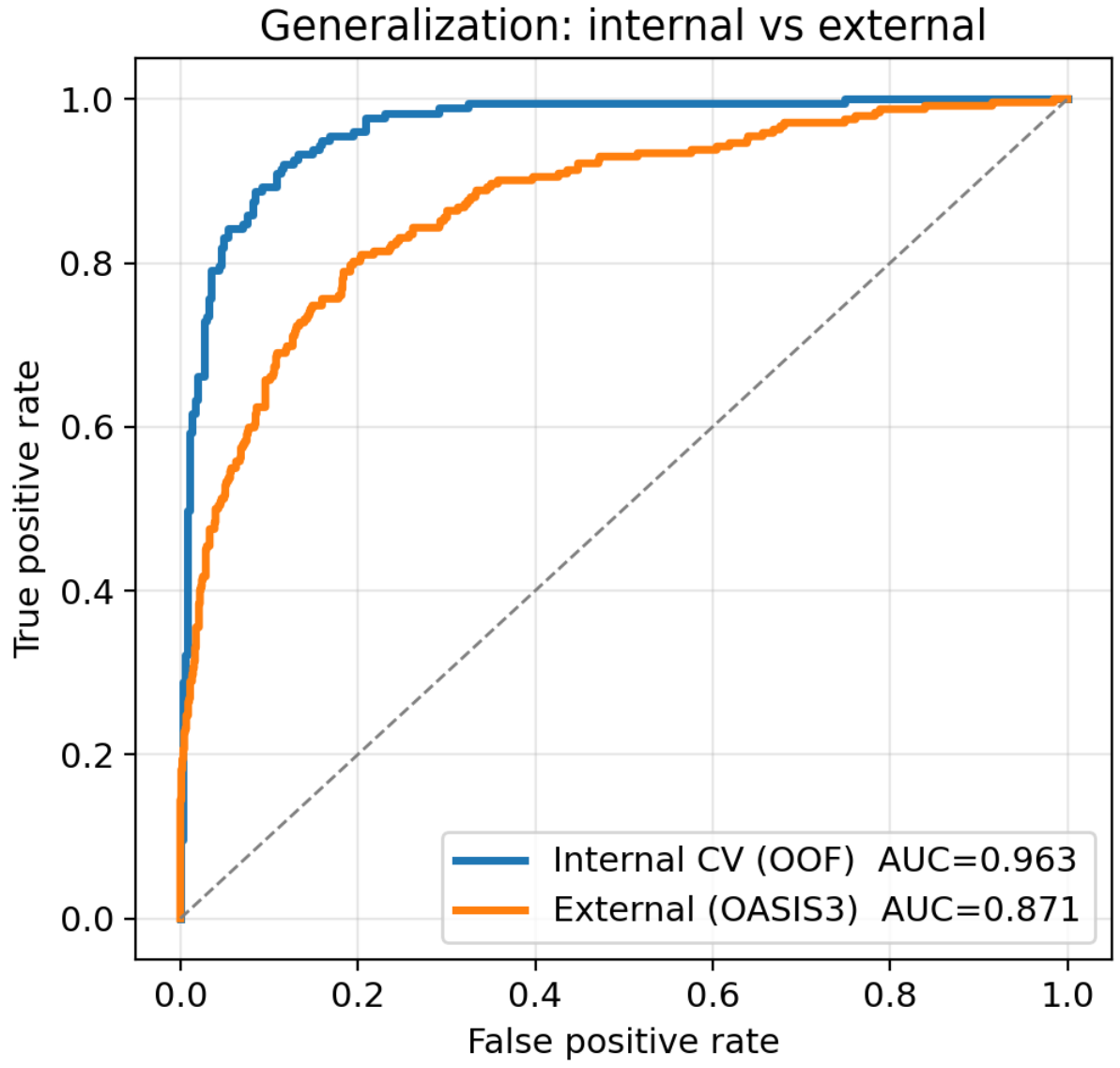

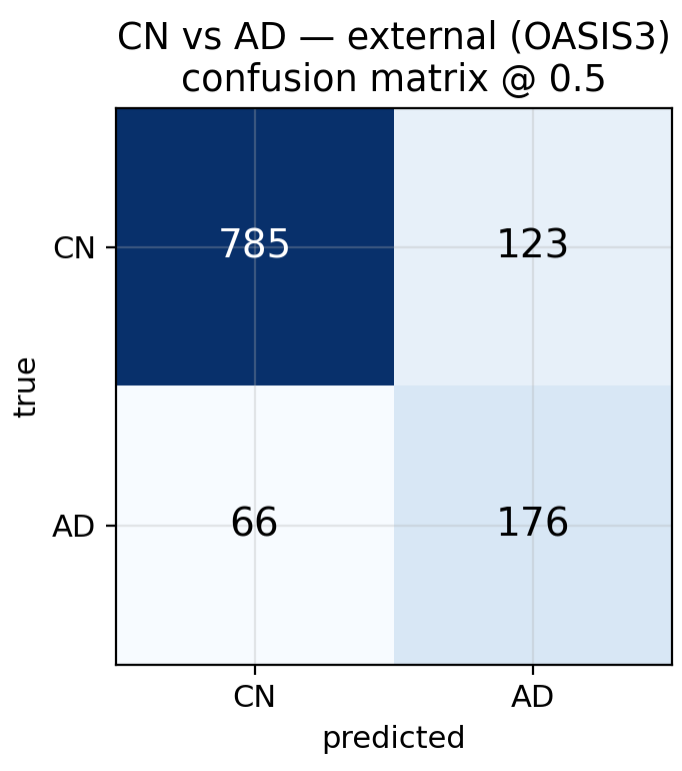

**Figure 3.1** ROC curve for Cognitively Normal vs Dementia classification for internal ADNI cohort (AUC = 0.964, 95% CI 0.947–0.978) vs the external OASIS-3 cohort (AUC = 0.871, 95% CI 0.844–0.896). The dashed diagonal marks chance.

**Figure 3.2** Confusion matrix on OASIS-3 at a 0.5 threshold (1,150 subjects, 242 with dementia), corresponding to a sensitivity of 0.727 and a specificity of 0.865.

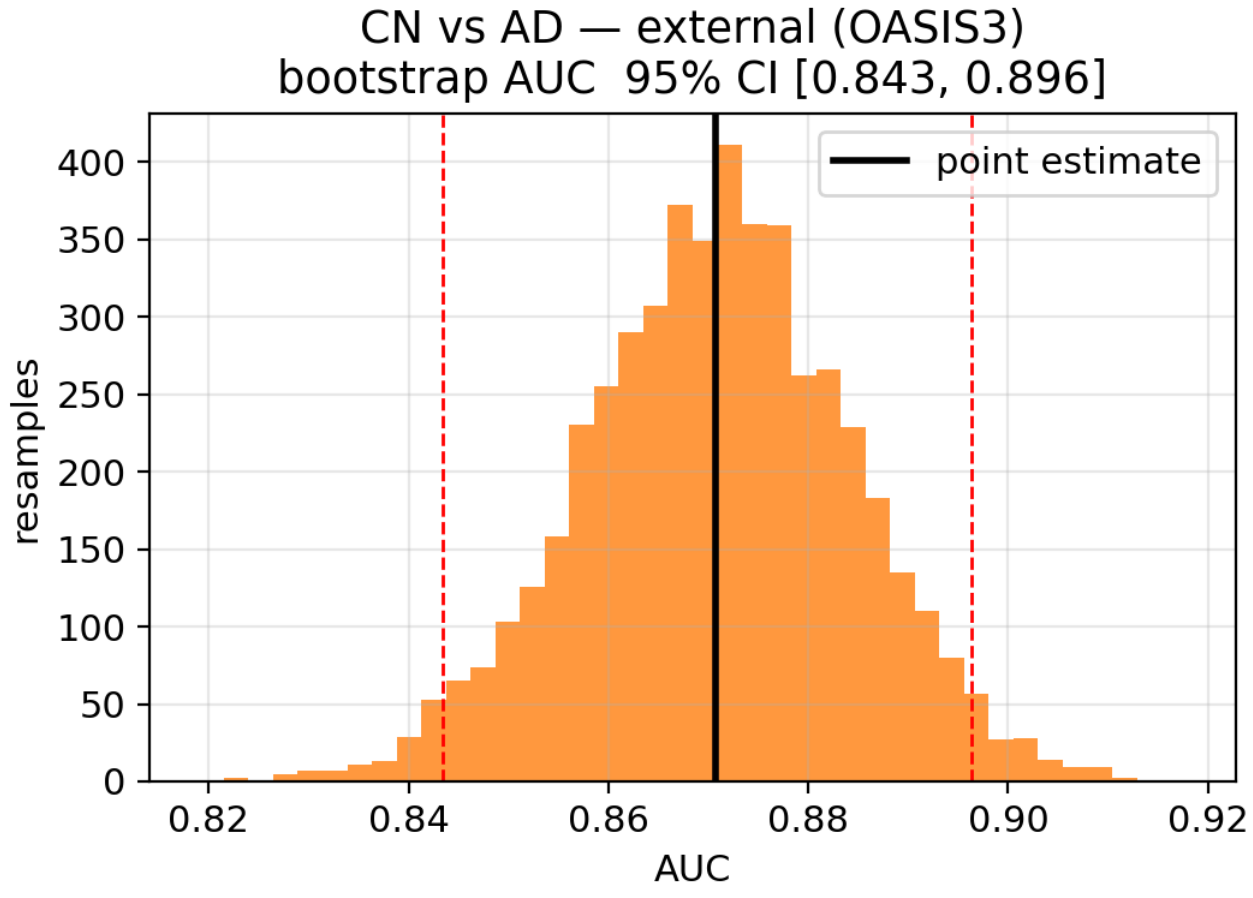

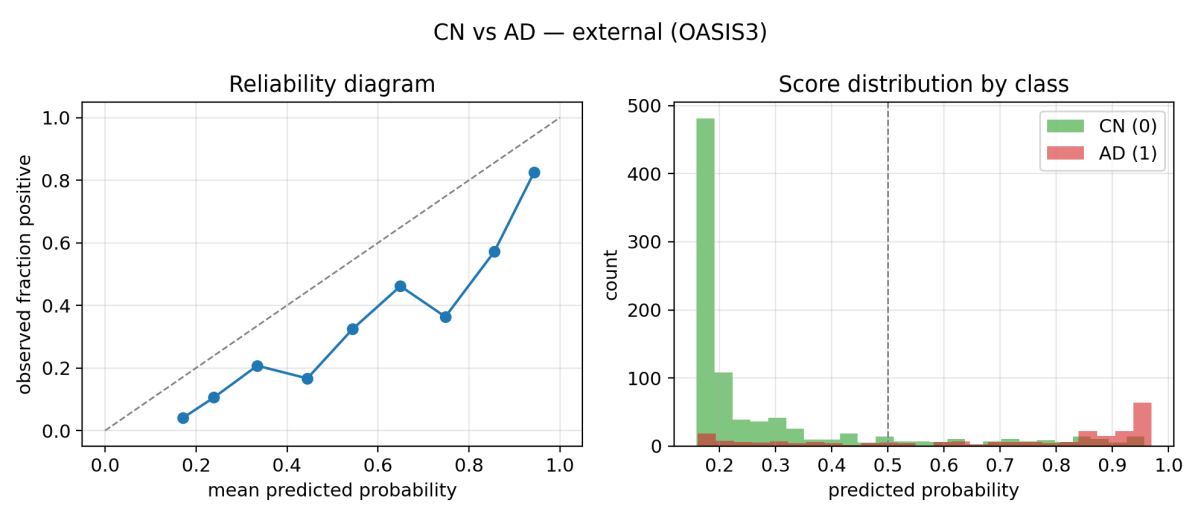

**Figure 3.3** Subject-bootstrap distribution of the external AUC (5,000 resamples).
The point estimate (0.871) and the 95% confidence interval (0.844–0.896) are indicated; the distribution remains clearly separated from chance on the unseen cohort.

**Figure 3.4** Calibration of the classifier on OASIS-3.
Left: reliability diagram (predicted vs observed probability).
Right: predicted-probability distributions for the cognitively normal and dementia groups.

# Experiment #3

Combining the Experiment #1 imaging logit with age and the ADAS-Cog-13 score, a pooled cross-validated logistic regression predicted conversion from MCI to dementia within approximately two years with an AUC of 0.828 (95% CI 0.761–0.890; Figure 4.1) over 210 MCI subjects, of whom 55 converted.

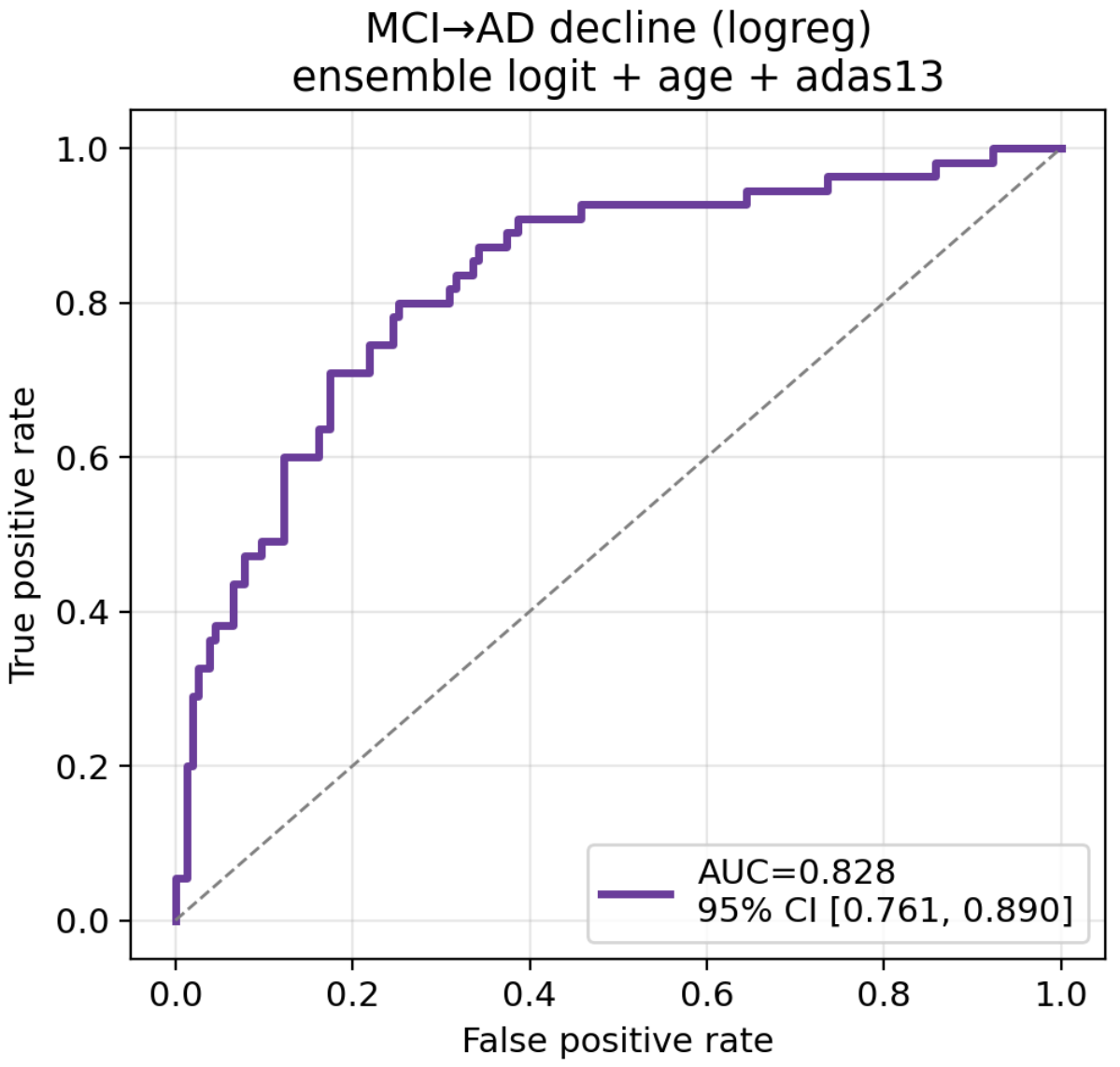

**Figure 4.1** Receiver operating characteristic (ROC) curve for predicting MCI-to-dementia conversion within ~2 years, using the Experiment #1 imaging logit together with age and ADAS-Cog-13 (pooled out-of-fold logistic regression; AUC = 0.828, 95% CI 0.761–0.890; 210 subjects, 55 converters). The dashed diagonal marks chance.

**Figure 4.2** Predicted conversion risk stratified by true outcome. Subjects who converted to dementia (n = 55) received markedly higher predicted risk than those who remained stable (n = 155); boxes show the interquartile range with individual subjects overlaid.

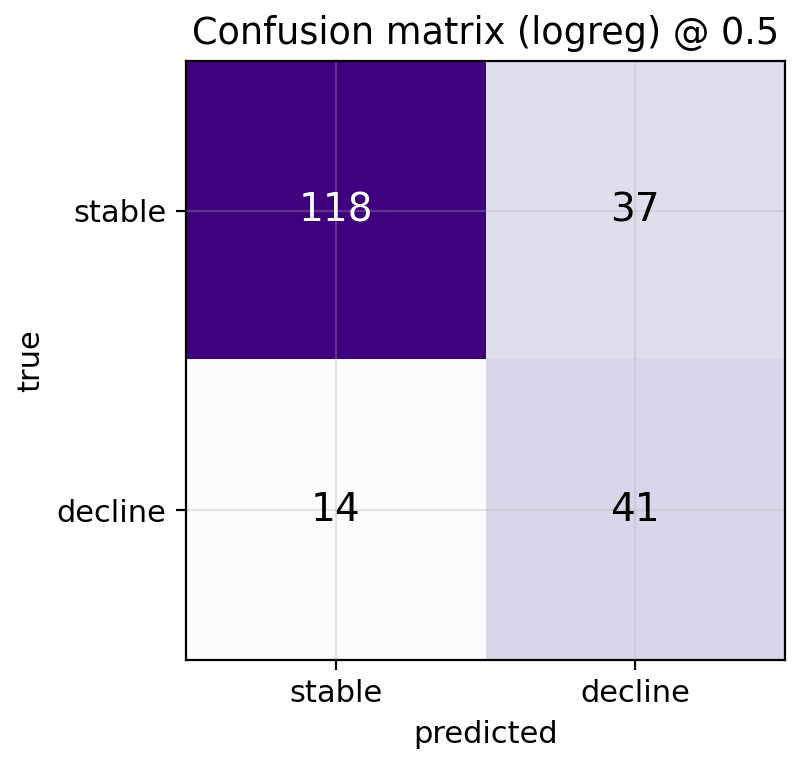

**Figure 4.3** Confusion matrix for MCI-to-dementia conversion at a 0.5 threshold
(210 subjects), corresponding to a sensitivity of 0.745 (41/55 converters detected) and a specificity of 0.761 (118/155 stable subjects correctly classified).

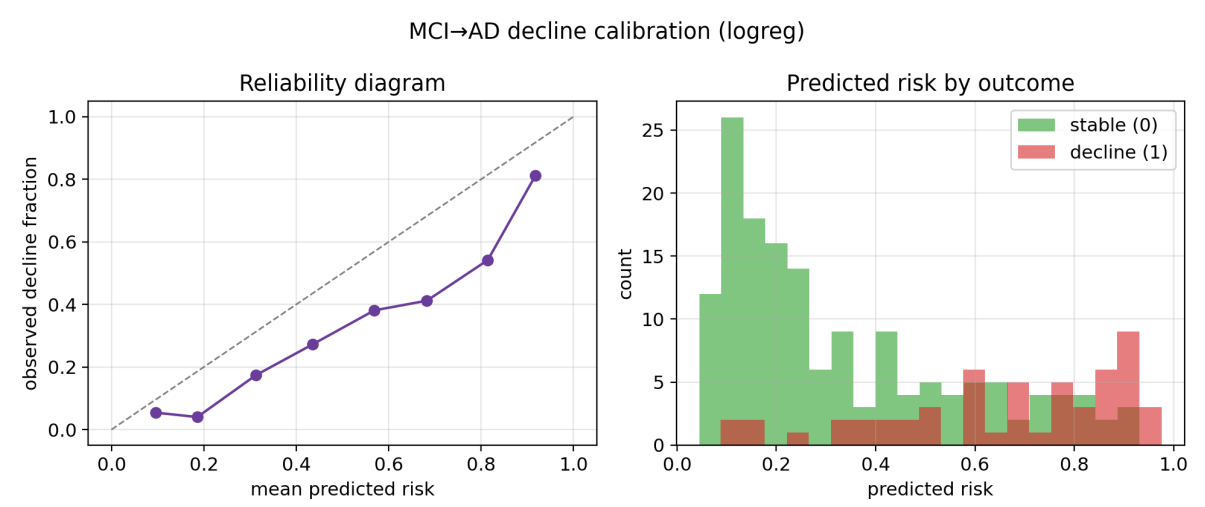

**Figure 4.4** Calibration of the conversion model.
Left: reliability diagram (predicted vs observed conversion rate).
Right: predicted-risk distributions for stable and converting subjects.

Predicted conversion risk was substantially higher in subjects who went on to convert than in those who remained stable (Figure 4.2). At a 0.5 decision threshold the model achieved a balanced accuracy of 0.753, a sensitivity of 0.745 (95% CI 0.627–0.857), and a specificity of 0.761 (95% CI 0.694–0.825), correctly identifying 41 of the 55 future converters while misclassifying 37 of the 155 stable subjects (Figure 4.3). The predicted probabilities were reasonably well calibrated across the risk range (Figure 4.4).

## Experiment #4

We estimated amyloid positivity using only structural MRI data, without including any other features. The fine-tuned model achieved a pooled out-of-fold AUC of 0.804 (95% CI 0.775–0.832; Figure 5.1) across 865 subjects, with 405 of them being amyloid-positive. Using a decision threshold of 0.5, the model reached a balanced accuracy of 0.734, a sensitivity of 0.738 (95% CI 0.696–0.779), and a specificity of 0.730 (95% CI 0.688–0.770). It correctly identified 299 out of 405 amyloid-positive subjects, while producing 124 false positives among the 460 amyloid-negative subjects (Figure 5.2). The subject-bootstrap distribution of the AUC was clearly above chance (Figure 5.3). The predicted probabilities were reasonably calibrated, but the score distributions for the two groups overlapped considerably, which shows the difficulty of this task (Figure 5.4). Training was stable in all folds, but the two losses did not follow the same course. Training loss decreased monotonically, whereas validation loss reached its minimum at around the fourth epoch and then drifted slightly upward, so that the curves separated over the remainder of training (Figure 5.5). Because we selected the checkpoint at minimum validation loss, the divergence in later epochs does not enter the reported results. Although the performance was lower than for distinguishing CN from Dementia, the model still showed above-chance discrimination. This suggests that the encoder is able to capture structural features related to amyloid burden.

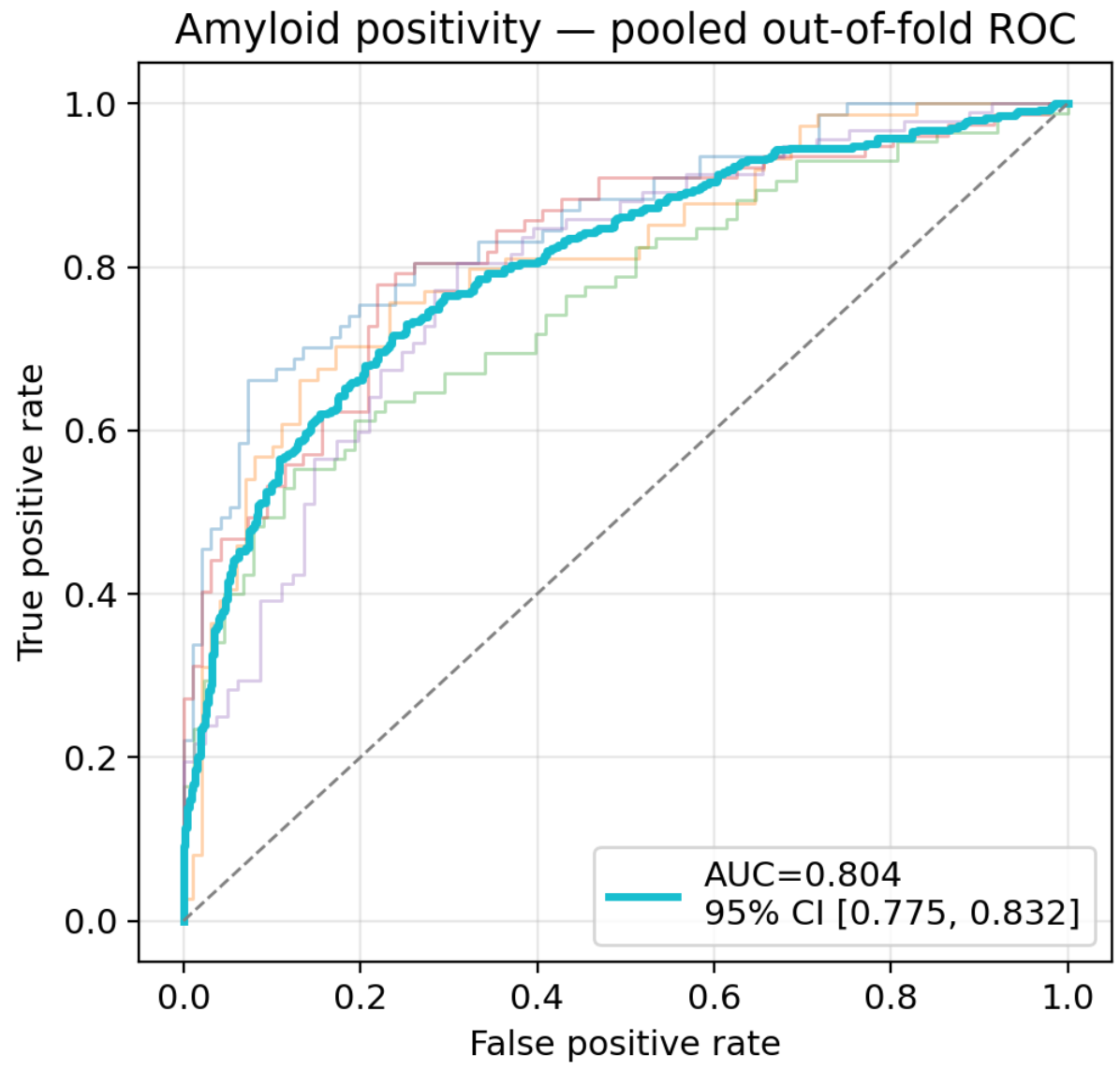

**Figure 5.1** ROC curve for amyloid positivity (Aβ− vs Aβ+) from structural MRI alone (pooled out-of-fold ensemble; AUC = 0.804, 95% CI 0.775–0.832; 865 subjects, 405 amyloid-positive). Thin lines show the five individual folds; the dashed diagonal marks chance.

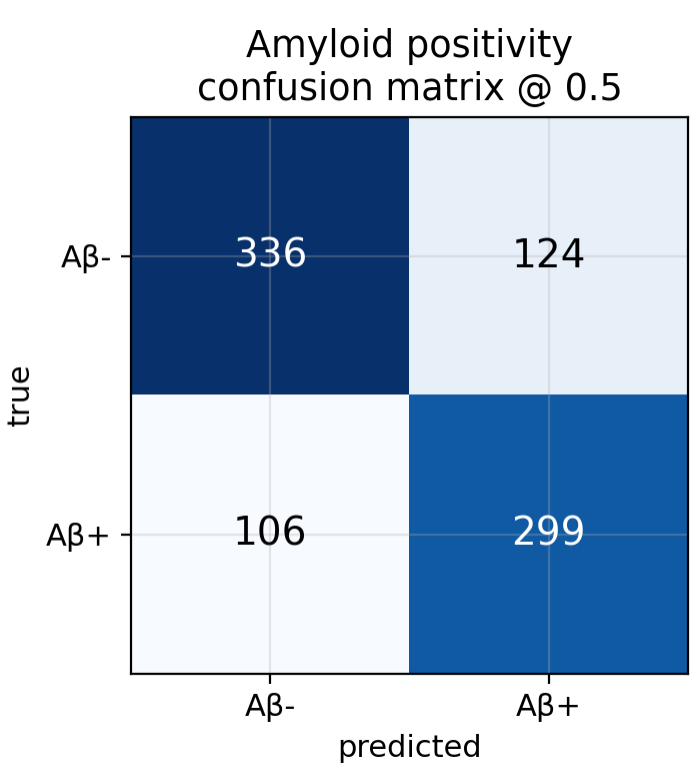

**Figure 5.2** Confusion matrix for amyloid positivity at a 0.5 threshold (865 subjects; 405 Aβ+, 460 Aβ−), corresponding to a sensitivity of 0.738 (299/405 detected) and a specificity of 0.730 (336/460 correctly classified).

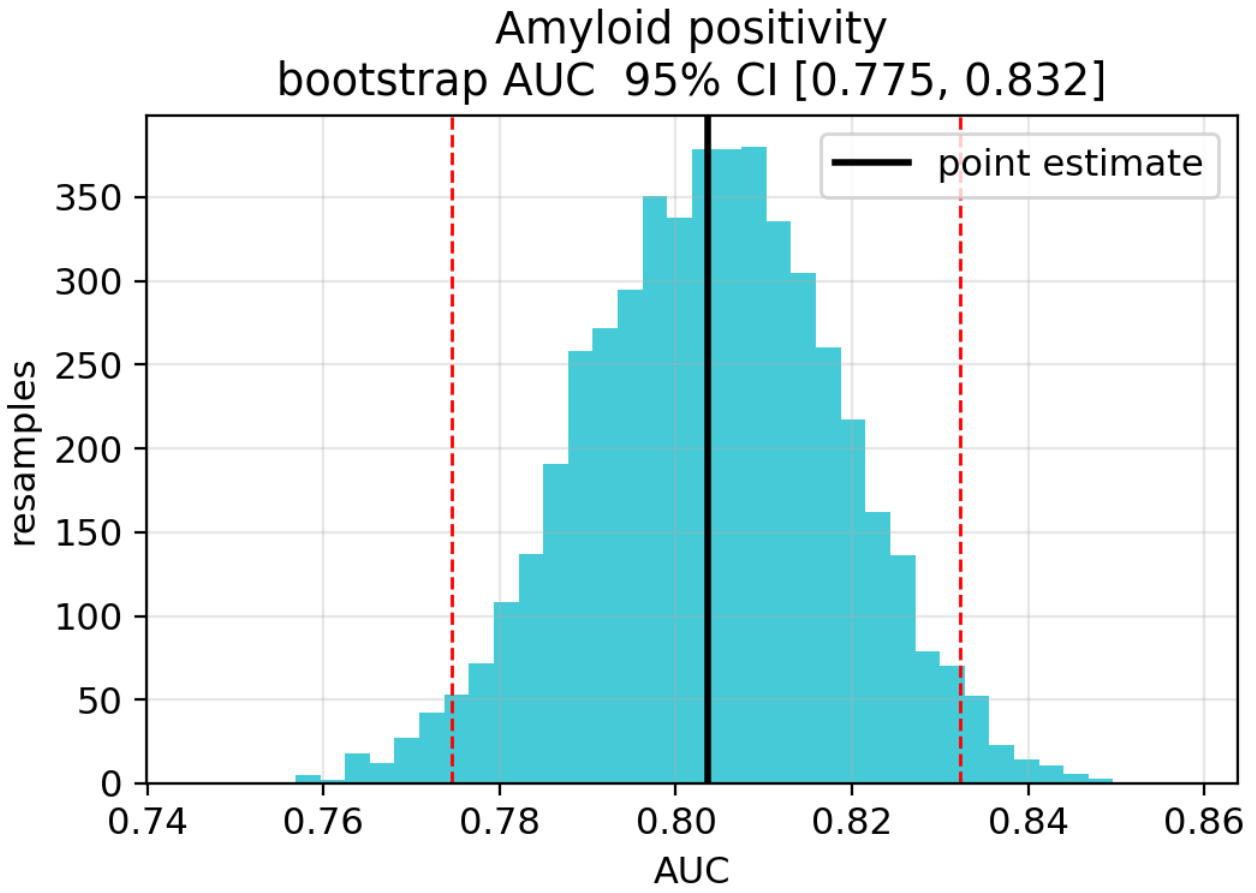

**Figure 5.3** Subject-bootstrap distribution of the amyloid-positivity AUC (5,000 resamples). The point estimate (0.804) and the 95% confidence interval (0.775–0.832) are indicated; the distribution is separated from chance.

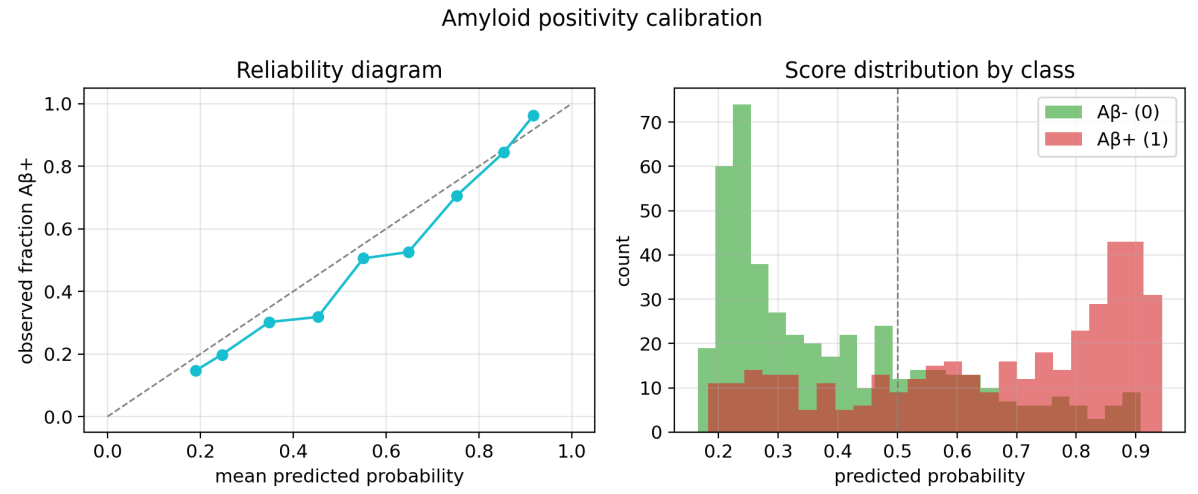

**Figure 5.4** Calibration of the amyloid classifier.
Left: reliability diagram (predicted vs observed positivity rate). Right: predicted-probability distributions for amyloid-negative and amyloid-positive subjects, which overlap substantially, reflecting the difficulty of inferring amyloid status from structural MRI.

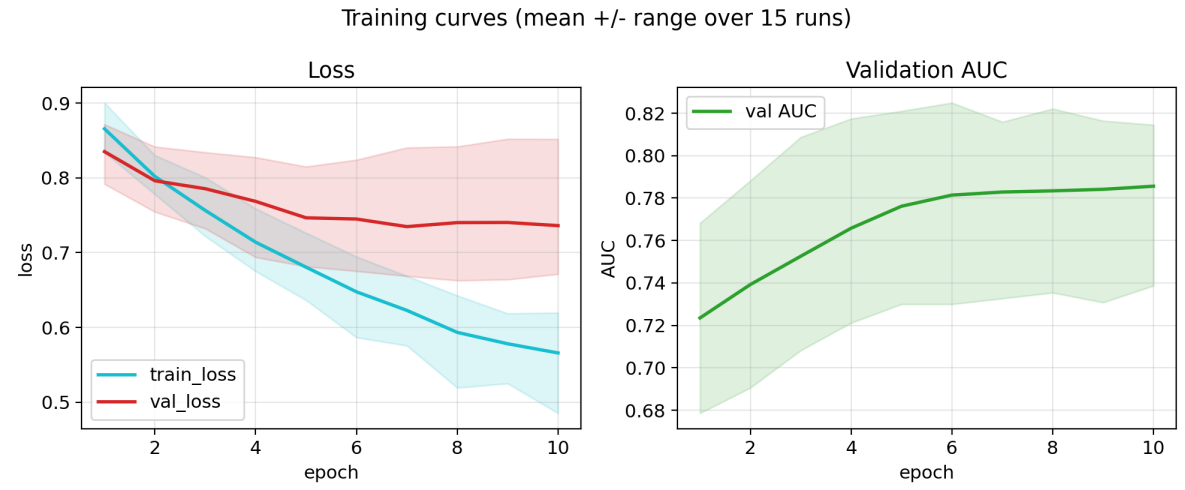

**Figure 5.5** Training dynamics across the 15 fold × seed runs (mean ± range).
Left: training loss falls throughout; validation loss reaches a minimum near epoch 4, then rises by less than the spread across runs. Right: validation AUC rises to a plateau and does not decline. Reported results use the checkpoint at minimum validation loss.

# Experiment #5

We evaluated whether the encoder can learn features specific to the hippocampus by predicting ICV-normalized hippocampal volume directly from MR images. Using cross-validation with 5 folds and 3 random seeds on 2,092 subjects, the model's predictions were very similar to the measurements from FreeSurfer. The pooled out-of-fold $R^2$ was 0.797 (95% CI 0.769 to 0.821), and the mean absolute error was $2.6 \times 10^{-4}$ (see Figure 6.1). The residuals were close to zero and did not show any clear pattern across the range of measurements, which suggests that the predictions were unbiased (Figure 6.2). The distribution of $R^2$ values from bootstrapping was also narrow (Figure 6.3). Training was stable in all runs, with validation $R^2$ reaching a plateau as the loss decreased. The $R^2$ for each fold ranged from 0.73 to 0.84.

Figure 6.1 Predicted versus measured ICV-normalized hippocampal volume
(pooled out-of-fold predictions; 2,092 subjects across all five folds). Points lie close to the identity line (dashed), with $R^2$ = 0.797 (95% CI 0.769–0.821) and a mean absolute error of $2.6 \times 10^{-4}$.

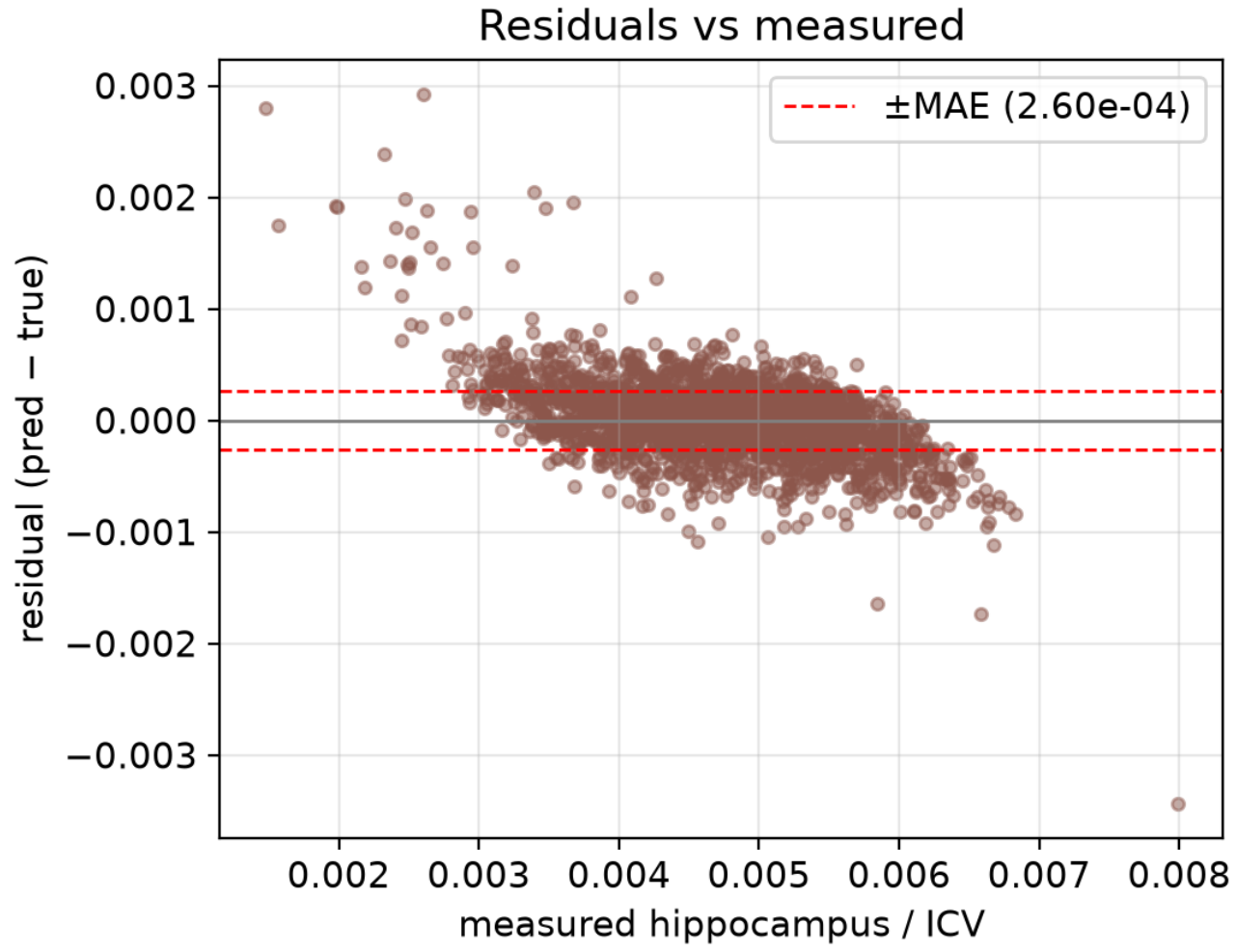

**Figure 6.2** Prediction residuals (predicted − measured) versus measured ICV-normalized hippocampal volume.
Residuals are centred near zero (grey line) with no systematic trend across the range; the red dashed bands mark ± the mean absolute error.

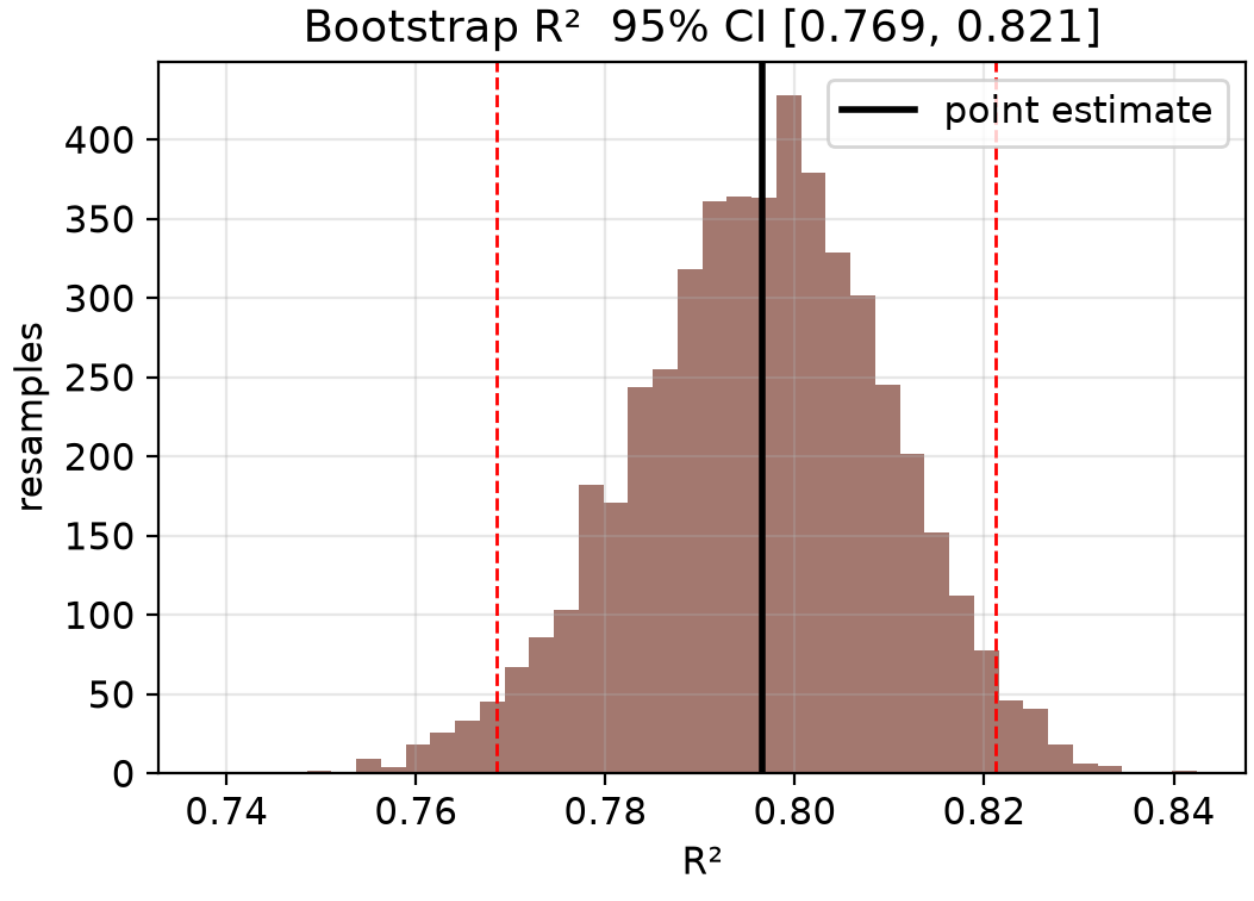

**Figure 6.3** Subject-bootstrap distribution of the pooled $R^2$ (5,000 resamples).
The point estimate (0.797) and the 95% confidence interval (0.769–0.821) are indicated.

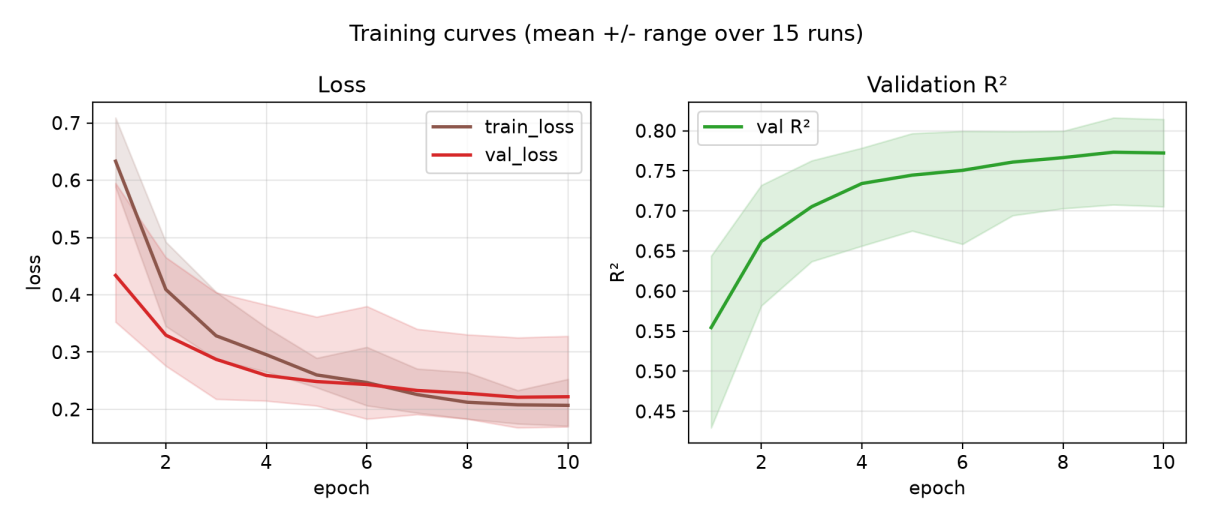

**Figure 6.4** Training dynamics across the 15 fold × seed runs (mean ± range).
Left: training and validation loss versus epoch. Right: validation $R^2$ versus epoch, which plateaus as the losses converge, indicating stable fine-tuning.

# Experiment #6

Our second continuous target was the volume of white matter hypointensities, normalized by ICV, which we predicted from the same MR images. We evaluated the model using 5-fold cross-validation with 3 different random seeds, covering 2,092 subjects. The pooled out-of-fold $R^2$ was 0.913 (95% CI 0.893 to 0.931), and the mean absolute error was $6.0 \times 10^{-4}$ (Figure 7.1). The $R^2$ values for each fold were between 0.90 and 0.92. The residuals were close to zero and did not show any pattern across the predicted range (Figure 7.2). The distribution of $R^2$ values from subject bootstrapping was narrow, indicating stable performance (Figure 7.3). In all 15 runs, the validation $R^2$ reached a plateau as the loss values decreased (Figure 7.4).

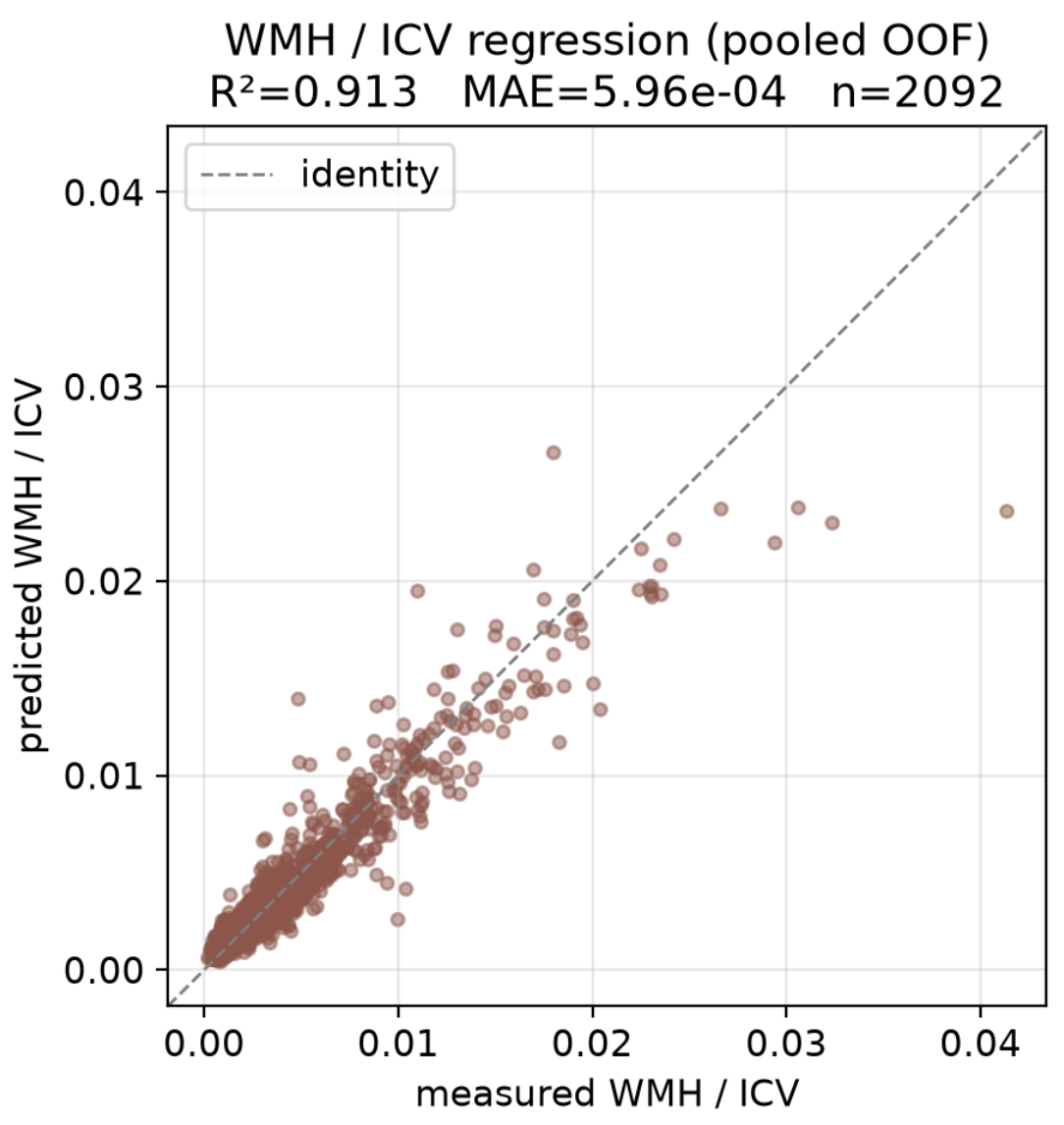

**Figure 7.1** Predicted versus measured ICV-normalized white matter hypointensity volume (pooled out-of-fold predictions; 2,092 subjects across all five folds). Points lie close to the identity line (dashed), with $R^2 = 0.913$ (95% CI 0.893–0.931) and a mean absolute error of $6.0 \times 10^{-4}$.

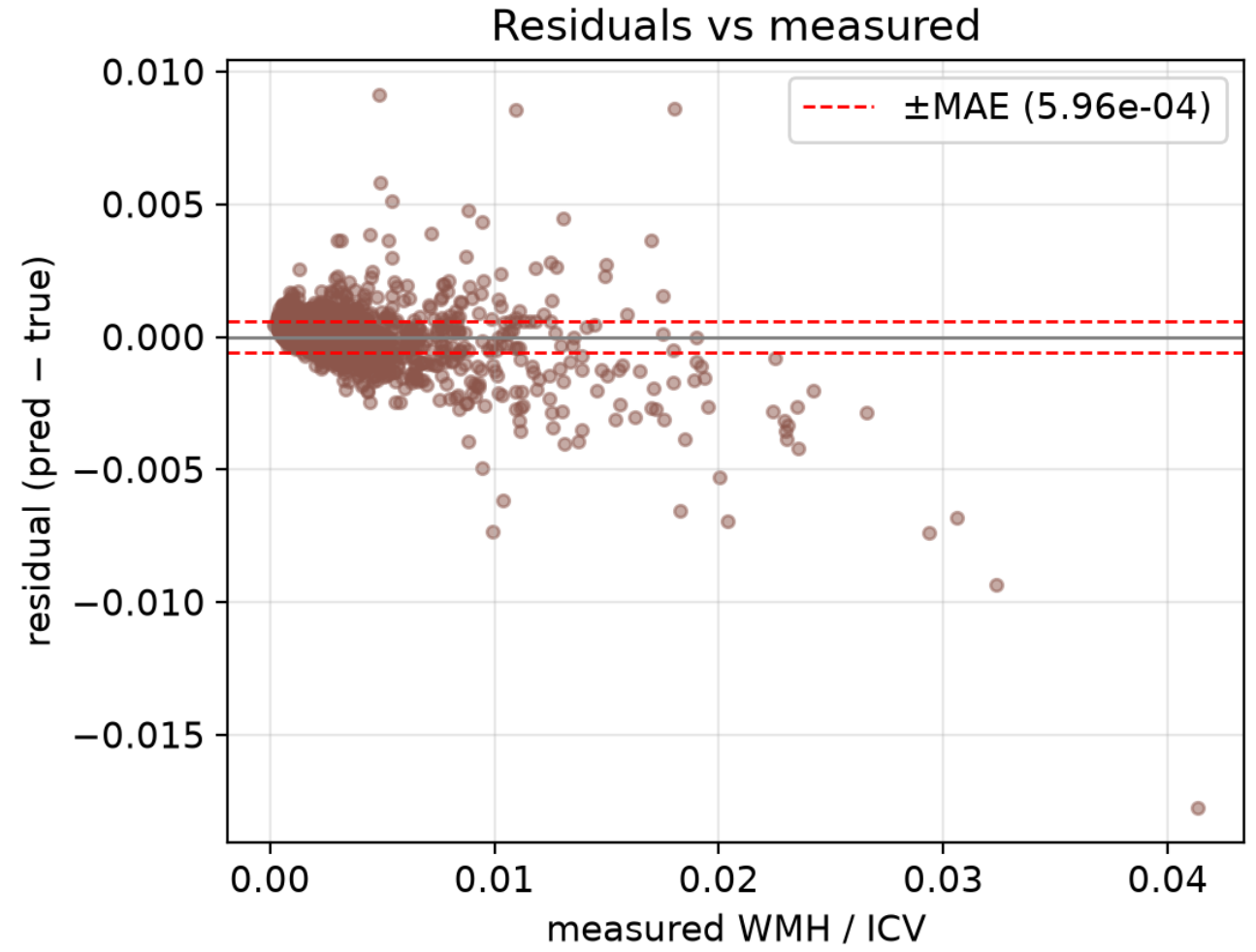

**Figure 7.2** Prediction residuals (predicted − measured) versus measured white matter hypointensity volume normalized by ICV. Residuals are centred near zero (grey line) with no systematic trend across the range; the red dashed bands mark ± the mean absolute error.

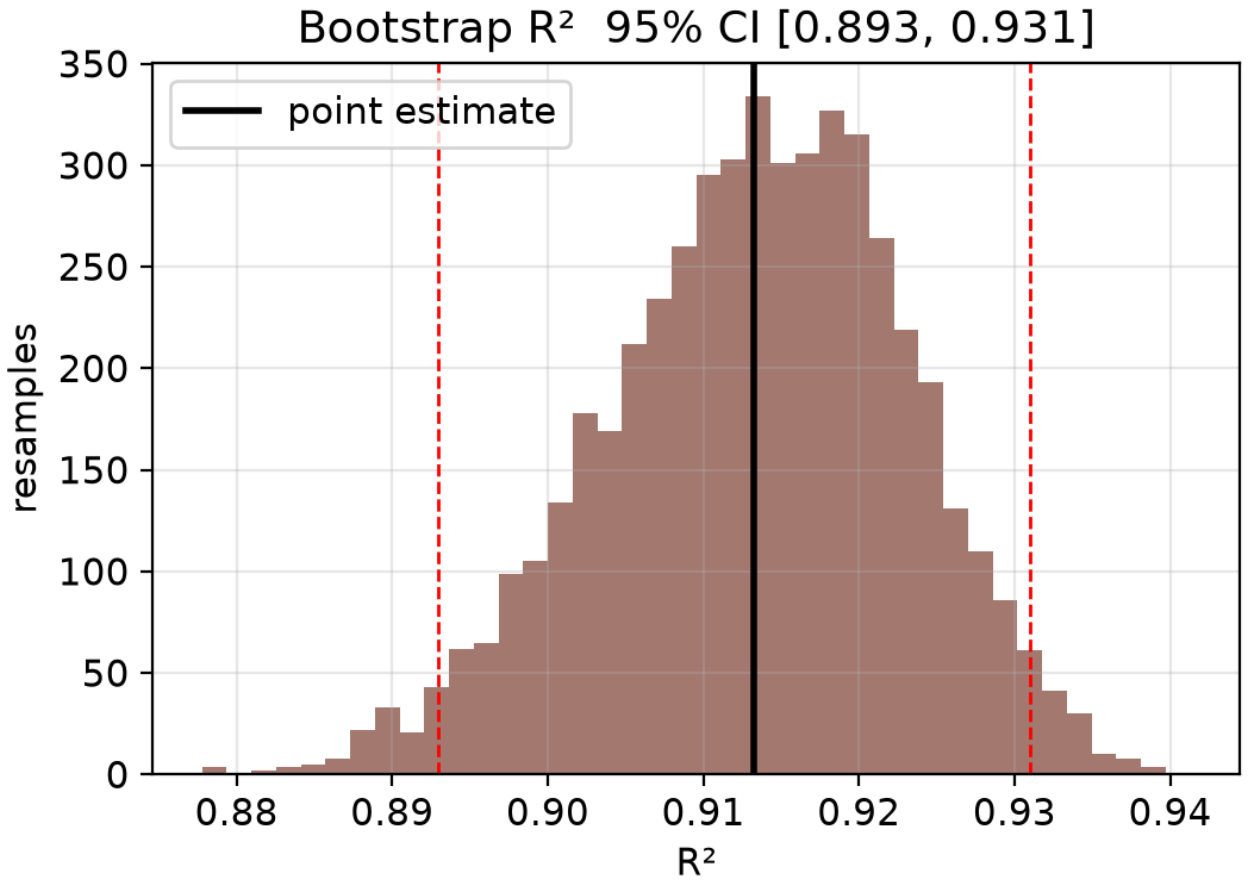

**Figure 7.3** Subject-bootstrap distribution of the pooled $R^2$ (5,000 resamples).
The point estimate (0.913) and the 95% confidence interval (0.893–0.931) are indicated.

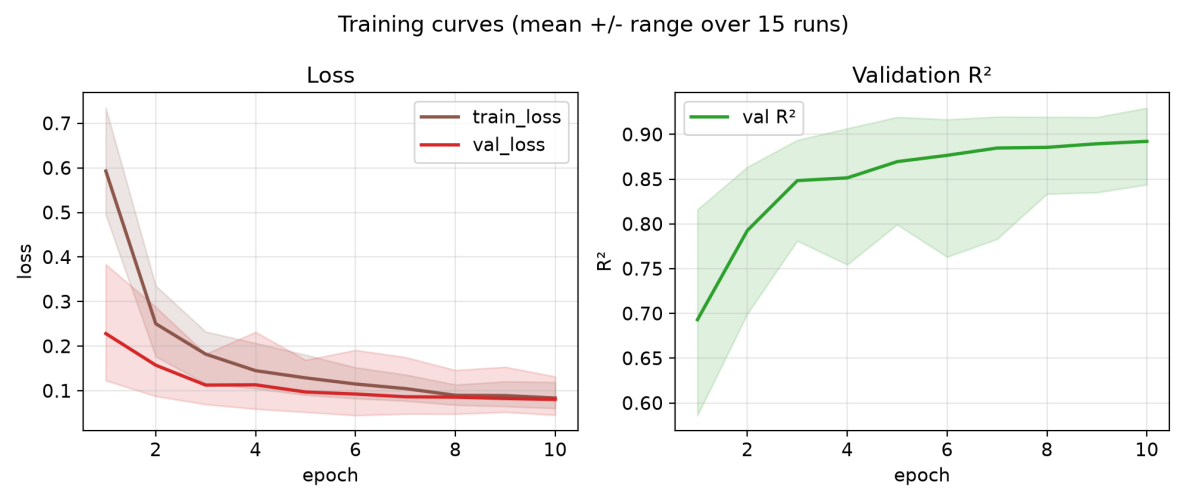

**Figure 7.4** Training dynamics across the 15 fold × seed runs (mean ± range).
Left: training and validation loss versus epoch. Right: validation $R^2$ versus epoch, which plateaus as the losses converge, indicating stable fine-tuning.

# Discussion

Across six experiments, a brain age model kept frozen and adapted with less than 1% new trainable parameters performed comparably to previously reported results obtained with larger models or hand-engineered features. This supports treating it as a generalizable feature extractor for Alzheimer's-related tasks. This matters because MRI with age labels is abundant, while MRI labelled for a specific disease is not.

On held-out ADNI folds our model reached an AUC of 0.964 and a balanced accuracy of 0.889, using no clinical variables as input. Experiments #2 and #3 build on this adapted model.

When applied to OASIS-3 without retraining, recalibration, or adaptation, the same model reached an AUC of 0.871, a drop of about 0.09 from the within distribution result. This is the result we consider most important, because existing literature does not often report results for within distribution and out-of-distribution cohorts. AnatCL [6] and Dufumier et al. [5] both fit a new classifier on each target dataset, so their protocol cannot produce a result for transfer without target-side fitting. The study that does report out-of-distribution performance without target-side fitting is weaker: Mazher et al. [1] documented an AUC of 0.77, 0.71, and 0.73 on NACC, OASIS, and AIBL. Bron et al. [4] retained an AUC of 0.90 on an external cohort, but with a model built for that single task rather than a reusable encoder. We note that these comparisons involve different external cohorts, so they indicate the general level of performance rather than a direct ranking.

The two cohorts (ADNI and OASIS-3) are similar in sex distribution. Among controls, 58.4% are female in one cohort and 57.8% in the other. Among those with dementia, the percentages are 45.9% and 46.9%. However, the cohorts differ in age. On average, OASIS-3 controls are 4.7 years younger and show more variation in age. This difference in age could make it easier for a brain age model to perform well on the external task. The cohorts also differ in how sharply their diagnostic groups are separated: APOE ε4 carriage prevalence differs by 45 percentage points between controls and dementia cases in ADNI (27.3% vs 71.8%) but by only 28 in OASIS-3 (35.3% vs 63.6%). This suggests that OASIS-3 is less sharply selected. The observed decrease in performance is likely due to both differences in data acquisition, and the less selected, younger population. This situation is similar to what a model would face in real-world use.

Using the Experiment #1 logit together with age and ADAS-Cog-13, we predicted conversion from MCI to dementia within two years with a balanced accuracy of 0.753. Shafiee et al. [17] reported 0.759 ± 0.011 on the same cohort using age, cognitive scores, and SNIPE scores for the hippocampus and entorhinal cortex. Our result is equivalent, not better. What changes is the amount of work required to get it: the SNIPE score needs both EC and HC structures to be segmented for every participant and a lookup library of CN and dementia cases, whereas here the imaging contribution is a single number produced by a model that was never trained on MCI. The encoder had seen no ADNI MCI scans at any point, so the transfer is genuinely to a new population as well as a new task.

From structural T1w MRI alone, with no covariates, the model classified amyloid positivity with an AUC of 0.804. This sits in the upper part of the published range for T1w-only models, which spans AUCs of 0.610 [22], 0.730 [20], and 0.857 [21]. Only the last exceeds ours, and it comes from a vision transformer trained end-to-end for this task. The score distributions of the two groups overlap substantially, which is expected, since amyloid deposition is not directly visible on T1w images. The result is still clearly above chance, which indicates that the frozen representation carries structural signals associated with amyloid burden even though nothing in the brain age pretraining task relates to molecular pathology.

We estimated ICV-normalized hippocampal volume directly from the images with an $R^2$ of 0.797 across 2,092 participants. Because the model operates on images registered to the MNI stereotaxic template [24], [25], head size is already normalized out, so the quantity the network recovers is a normalized volume rather than a raw one. This is the more useful target, since normalized volumes are what group-level analyses use, but it is also the harder one, because normalization removes variance associated with head size that would otherwise make the prediction easier. Most published work on this measurement uses U-Net segmentation architectures, typically around 20 million parameters, trained end to end. That a frozen encoder supports the same readout with about 1% new parameters suggests either that the representation retains localized structural information, or that normalized hippocampal volume is partly recoverable from more global patterns. Distinguishing the two would require an occlusion or saliency analysis, which we have not carried out. Grad-CAM maps computed on the same encoder family in [7] were spatially localized and included medial temporal regions, which is consistent with the first explanation.

The same model architecture estimated ICV-normalized white matter hypointensity volume with an $R^2$ of 0.913. Both of these reference measures are produced by FreeSurfer from the same T1w images our model receives, so these two experiments ask whether the frozen representation retains enough structural detail to reproduce the output of a dedicated morphometric pipeline. It does, for two very different targets: one small, bilateral, and anatomically fixed, the other diffuse and variable in location. Neither target has any relationship to the age-prediction objective the encoder was trained on. This suggests that brain age pretraining compresses the image not only into a global summary of aging, but into a latent representation that also preserves local structural detail, enough to support measurements the model was never trained to make.

Our encoder is a 6-block residual 3D CNN with 7.18 million parameters. AnatCL [6] uses a 3D ResNet-18 and Dufumier et al. [5] a 3D DenseNet121; neither paper reports a parameter count, but both architectures are considerably larger than ours. On CN versus AD dementia in ADNI, Bron et al. [4] report a maximum AUC of 0.94 with models trained specifically for that task, Mazher et al. [1] report 0.81 with a general-purpose brain foundation model, and Dufumier et al. report a fine-tuned AUC of 0.968. Our 0.964 is therefore equivalent to the best of these rather than better. Measured by balanced accuracy, AnatCL reports 0.808 against our 0.889. On OASIS-3 it reports 0.787 against our 0.796, and here the protocols differ in our disfavour: AnatCL fits a linear classifier on OASIS-3, while our model never sees that dataset at all. We do not compare our sMCI versus pMCI result with theirs, because ours combines the imaging logit with age and ADAS-Cog-13 while theirs uses imaging alone; the appropriate comparison there is with Shafiee et al. [17], who used the same features. These comparisons are indicative rather than exact, since the cohorts, the preprocessing (AnatCL uses VBM with CAT12) and the MCI definitions all differ. One difference runs against us and should be stated plainly: our encoder was pretrained on 26,512 scans, while AnatCL used 3,984 and Dufumier et al. about 10,000. Our claim is therefore about model capacity rather than the volume of pretraining data.

A recent preprint makes a similar claim. BrainDINO [26] is a self-supervised model pretrained on about 6.6 million two-dimensional slices from 20 datasets. It keeps the encoder frozen and trains only a small head, updating as little as 0.6% of the parameters, and reports a macro-AUC of 0.850 for CN versus AD on ADNI. Our approach differs in three ways. Ours is pretrained on whole volumes rather than single slices, and on chronological age rather than a self-supervised objective. More importantly, their OASIS evaluation trains on 195 subjects and tests on 40 from the same cohort, while we apply an unchanged model to 1,150 OASIS-3 participants and fit nothing on that dataset.

Our results point to two suggestions for future work on brain MRI foundation models. First, an age-supervised model is worth including as a baseline. Two of the studies closest to ours already contain one, and in both it performs well, but neither pursues it further. Second, we would encourage reporting performance on an external cohort without any target-side fitting, alongside cross-validation results. It describes what a new site would obtain, and it is rarely reported. Third, a compact encoder adapted with a small number of added parameters is worth trying before a larger one. Our model reached this level of performance with 7.18 million frozen parameters and 72,249 per task, and adding a task cost an adapter rather than a new network.

# Limitations

First, we did not tune hyperparameters; as noted above, this makes the reported results a conservative lower bound. Second, white matter hypointensity volumes are strongly right-skewed, with most participants showing little burden and a small number showing a great deal. The $R^2$ we report for that task is therefore influenced by those few participants and should not be read as directly comparable to the $R^2$ for hippocampal volume, which has a much more symmetric distribution. Finally, and most importantly for how these results should be read, every downstream task we tested lies on the Alzheimer's continuum, and the pretraining task, brain age prediction, plausibly encodes much the same atrophy patterns that those tasks depend on. Our results therefore support using a brain age encoder for Alzheimer's-related problems; they do not establish that it would transfer to conditions with different structural signatures, such as tumors, demyelination, or psychiatric disorders. There is indirect evidence that age-supervised pretraining helps beyond this continuum, since Dufumier et al. [5] report it to be competitive for schizophrenia and bipolar disorder classification, but we have not tested that ourselves.

# Conclusions

Taken together, our results support a simple claim: a compact model trained with supervision on brain age is a usable base model for Alzheimer's-related tasks, and it is usable under the strict conditions that matter in practice, with the encoder frozen, a handful of new parameters, small target datasets, and in one case no target data at all. Including the different sets of LoRA weights we used, our model has fewer than 7.5 million parameters. Both of the closest existing works contain an age-supervised baseline that performs well, but neither pursues it. Our results suggest it deserves to be pursued. This approach can also be used when training a brain foundation model.